\documentclass[manuscript]{acmart}
\usepackage{soul} 
\usepackage{color, xcolor} 
\usepackage{xcolor}
\makeatletter
\def\bstctlcite{\@ifnextchar[{\@bstctlcite}{\@bstctlcite[@auxout]}}
\def\@bstctlcite[#1]#2{\@bsphack
	\@for\@citeb:=#2\do{%
		\edef\@citeb{\expandafter\@firstofone\@citeb}%
		\if@filesw\immediate\write\csname #1\endcsname{\string\citation{\@citeb}}\fi}%
	\@esphack}
\makeatother
\usepackage{threeparttable}
\usepackage{pifont}
\usepackage{multirow}
\usepackage{nomencl}
\makenomenclature
\usepackage{etoolbox}
\renewcommand\nomgroup[1]{%
	\item[\bfseries
	\ifstrequal{#1}{D}{Micro-expression Dataset}{%
		\ifstrequal{#1}{M}{Main abbreviations}{%
			\ifstrequal{#1}{O}{Other symbols}{}}}%
	]}
\usepackage{mathtools}

\newcommand{\maes}{\mbox{MaEs}}
\newcommand{\mies}{\mbox{MiEs}}

\begin{document}
	\title{Facial Expression Analysis and Its Potentials in IoT Systems: A Contemporary Survey}
	\author{Zixuan Shangguan}
	\affiliation{%
		\institution{Shenzhen MSU-BIT University} \city{Shenzhen} \state{Guangdong} \country{China}}
	
	\author{Yanjie Dong}
	\authornotemark[1]
	\affiliation{%
		\institution{Shenzhen MSU-BIT University} \city{Shenzhen} \state{Guangdong} \country{China}}
	
	\author{Song Guo}
	\affiliation{\institution{The Hong Kong University of Science and Technology} \city{Hong Kong} \country{China}}
	
	\author{Victor C. M. Leung}
	\affiliation{%
		\institution{Shenzhen MSU-BIT University} \city{Shenzhen} \state{Guangdong} \country{China}}
	
	\author{M. Jamal Deen}
	\affiliation{\institution{AI Atlas Inc} \city{Hamilton} \state{Ontario} \country{Canada}}
	
	\author{Xiping Hu}
	\affiliation{%
		\institution{Shenzhen MSU-BIT University} \city{Shenzhen} \state{Guangdong} \country{China}}
	\authornote{Yanjie Dong and Xiping Hu are corresponding authors.}
	\email{huxp@bit.edu.cn}

	\begin{abstract}
Facial expressions convey human emotions and can be categorized into macro-expressions (MaEs) and micro-expressions (MiEs) based on duration and intensity. While MaEs are voluntary and easily recognized, MiEs are involuntary, rapid, and can reveal concealed emotions. The integration of facial expression analysis with Internet-of-Thing (IoT) systems has significant potential across diverse scenarios.
IoT-enhanced MaE analysis enables real-time monitoring of patient emotions, facilitating improved mental health care in smart healthcare. Similarly, IoT-based MiE detection enhances surveillance accuracy and threat detection in smart security.
Our work aims to provide a comprehensive overview of research progress in facial expression analysis and explores its potential integration with IoT systems. We discuss the distinctions between our work and existing surveys, elaborate on advancements in MaE and MiE analysis techniques across various learning paradigms, and examine their potential applications in IoT.
We highlight challenges and future directions for the convergence of facial expression-based technologies and IoT systems, aiming to foster innovation in this domain. By presenting recent developments and practical applications, our work offers a systematic understanding of the ways of facial expression analysis to enhance IoT systems in healthcare, security, and beyond.
	\end{abstract}
	\maketitle

	\vspace{-0.2 cm}
	\section{Introduction}\vspace{-0.2 cm}
	Since its inception in 1999, Internet-of-Things (IoT) technology has seamlessly integrated into the fabric of modern society, driving key innovations in both civilian and industrial sectors.
	The widespread adoption of IoT technology generates skyrocketing amounts of daily data that were traditionally processed in clusters of cloud servers.
	Due to the ever-increasing privacy concerns, the information process is shifted from the cloud to the network edge. 
	The low access latency and high data security of edge computing enable new application domains, such as emotion detection, traffic surveillance, and healthcare.	
	Since facial expressions convey 55\% of emotional messages during the exchange of moods, thoughts, feelings, and mental states, they have become a medium for sensing emotions in front of facility screens \cite{mehrabian1967decoding}. 
	Recent advancements in IoT technologies have driven the widespread application of facial expression analysis in practical scenarios.
	More specifically, IoT devices can first collect raw facial expression data for pre-processing. 
	To fully leverage low access latency and high data security during the implementation of real-time privacy-preserving systems, the pre-processed features are then uploaded to edge devices for decision-making. 
	Facial expressions can be divided into macro-expressions (\maes) and micro-expressions (\mies) based on their duration and intensity. 
%	Two types of emotions within \maes~and \mies~are illustrated in Fig.\ref{fig2}. 
	In terms of duration, \maes~approximately last 0.5 to 4 seconds, while \mies~last less than 0.5 seconds \cite{shen2012effects}.
	The short duration of \mies~conveys subtle emotional message, which makes emotion perception from \mies~a challenging task without specialized techniques and training. The differences between \maes~and \mies~are as follows. 
	
	\textbf{\maes:} can be perceived during regular interactions. 
	The recognition accuracy of \maes~ can exceed 97\% in a controlled laboratory due to the spontaneity and noise immunity. 
	\maes~recognition can be applied in multiple scenarios, e.g., autonomous driving, psychological health assessment, and human-computer interaction; 
	\textbf{\mies:} are involuntary and rapid. 
	Due to the short duration and subtle emotional cues, the detection of \mies~ can be challenging with naked eyes. 
	It is reported that the average recognition accuracy of~\mies~is around 47\% after specialized training \cite{frank2009see}. 
	Nevertheless, MiEs can reveal the true emotions of a person since they are instinctive and cannot be concealed. 
	Numerous MiE applications in IoT systems are being explored, such as lie detection, criminal identification, and security control.

	The objective of facial expression analysis lies in detecting the emotions of humans from \maes~and~\mies. 
	The research works on MaE analysis are mainly based on deep models that detect human emotions by leveraging the isolated static images and sequential dynamic images of \maes. 
	More specifically, the deep static MaE analysis performs well for two scenarios: (1) when several isolated images are available; and (2) when real-time expression recognition is required. 
	The deep dynamic MaE analysis can benefit from the comprehensive understanding of temporal dynamics of \maes. 
	Since the emotion recognition of \mies~is challenging due to the short duration and subtle emotional cues, the research works on holographic MiE analysis consist of two parts, i.e., MiE spotting and MiE recognition. 
	The MiE spotting aims at detecting \mies~within a given video, and the MiE recognition then uses the detected \mies~for MiE classification within a set of MiE categories. 
	Different from the MaE analysis, the MiE analysis can provide valuable insights into potentially hidden human emotions. 

	\vspace{-0.5 cm}
	\section{Recent Surveys on Facial Expression Analysis}\label{review}\vspace{-0.1 cm}
	Facial expression analysis is a critical tool for understanding human emotions and intentions. 
	The research on facial expression analysis started in the 1970s \cite{Ekman1978, Ekman1971}.
	Over the years of development, facial expression analysis has penetrated into our human daily lives. 
	More specifically, facial expression analysis can be categorized into MaE analysis and MiE analysis. 
	Therefore, we are motivated to review recent research contributions on both MaE and MiE analysis to better understand the current research stage.

	\vspace{-0.5 cm}
	\subsection{Recent Surveys on MaE Analysis}\vspace{-0.1 cm}
	MaE analysis is a burgeoning field at the intersection of computer vision and human emotion recognition and can offer transformative potential in healthcare, security, and human-computer interaction.
	Pioneering research has laid the groundwork for recognizing subtle expressions. 
	For example, Pantic et al. \cite{895976} introduced three key problems in automatic MaE analysis: (1) detection of an image segment as a face; (2) extraction of facial expression information; and (3) classification of the expression. 
	Yet, significant challenges remain in achieving robust, scalable, and generalizable solutions to MaE Analysis.
	Leveraging the power of deep learning, researchers endeavor to address these challenges, advancing the field through innovative algorithms and practical IoT applications.
	In \cite{li2020deep}, Li et al. provided a comprehensive review of MaE recognition based on deep learning.
	More specifically, the available datasets for MaE analysis and the principles of data selection and evaluation were explored \cite{li2020deep}. 
	Later, Canal et al. \cite{CANAL2022593} conducted an exhaustive systematic literature review encompassing $94$ leading methods from $51$ related articles. 
	Their review meticulously delineates the computational workflow for MaE recognition, such as, preprocessing, feature extraction, and classification algorithms. 
	Through a rigorous statistical analysis, Canal et al. \cite{CANAL2022593} revealed that traditional machine learning methods suffer from limited generalization capabilities though the high precision in classification can be achieved. 
	With the powerful feature capabilities, deep learning techniques thrive in scenarios involving larger and more diverse datasets.
	Karnati et al. \cite{10041168} introduced the cutting-edge models that are tailored for deep MaE recognition across diverse input modalities. 
	The comprehensive evaluations provide compelling evidence of enhanced performance and generalization potential of deep models that set new benchmarks for future research \cite{10041168}.

	Over the last decades, research on MaE analysis has undergone remarkable evolution that was transitioned from a formidable challenge to a structured and highly efficient process. 
	Once fraught with complexities, the MaE analysis workflow now adheres to a robust standard pipeline that encompasses preprocessing, feature extraction, and classification. 
	The integration of deep learning into this pipeline has been a transformative milestone, propelling algorithmic performance to exceed 97\% accuracy under controlled laboratory conditions—an unprecedented achievement.
	Despite these advancements, comprehensive overviews remain scarce to bridge the gap between foundational research and practical applications. 
	This limitation hampers the widespread adoption of MaE technologies in real-world (a.k.a., wild) scenarios, especially in areas like IoT integration and multimodal recognition.
	By bridging the gap between foundational research and wild application, we first introduce deep MaE recognition algorithms in Section \ref{MaErec} and the MaE applications based on IoT in Section \ref{maeiot}.
	Our analysis not only consolidates existing knowledge but also sets the stage for pioneering applications, ensuring our research addresses the pressing needs of this rapidly advancing field.

	\vspace{-0.5 cm}
	\subsection{Recent Surveys on MiE Analysis}\vspace{-0.1 cm}
	As MiE analysis continues to gain significant attention across academia and industry, numerous surveys have systematically reviewed advancements in MiE analysis from diverse perspectives. 
	For example, Ben et al. \cite{ben2021video} conducted the first comprehensive survey that offers a systematic examination of MiE analysis within a unified evaluation framework. 
	A neuropsychological perspective was provided on the distinction between MiEs and MaEs, i.e., the greater inhibitory effect of MiE on facial expressions in \cite{ben2021video}. 
	Additionally, several representative MiE datasets were also introduced and could be categorized into four distinct types: 
	(1) instructing participants to perform specific MiEs \cite{shreve2011macro,polikovsky2009facial}; 
	(2) constructing high-stakes scenarios \cite{warren2009detecting}; 
	(3) eliciting emotions through video stimuli while maintaining neutral expressions \cite{li2013spontaneous,yan2013casme,yan2014casme,davison2016samm,qu2017cas}; and 
	(4) capturing real high-stakes situations \cite{husak2017spotting}. 
	Notably, the micro-and-macro in expression warehouse (MMEW) dataset that contains high-resolution samples with a balanced distribution of MaE and MiE was proposed in \cite{ben2021video}. 
	Xie et al. \cite{xie2022overview} further advanced the understanding of MiE recognition by delving into critical areas, such as, macro-to-micro adaptation, recognition based on apex frames, and analysis leveraging facial action units (AUs). 
	Their detailed exploration offers a more nuanced perspective, complementing prior works and addressing emerging challenges in the field.
	Li et al. \cite{li2022deep} presented a pioneering survey on MiE analysis through the lens of deep learning techniques. 
	They systematically reviewed existing deep learning approaches by analyzing datasets, delineating the stepwise pipelines for MiE recognition, and conducting performance comparisons with state-of-the-art methods. 
	Their survey introduced a novel taxonomy for deep learning-based MiE recognition that can classify input data into static, dynamic, and hybrid categories. 
	Furthermore, Li et al. \cite{li2022deep} meticulously examined the architectural components of deep learning models, including network blocks, structural designs, training strategies, and loss functions. 
	Their findings emphasized that leveraging multiple spatiotemporal features as input yields superior performance in MiE recognition.
	Zhao et al. \cite{zhao2023facial} provided a holistic overview of MiE research from foundational psychological studies and early computer vision efforts to advanced computational MiE analysis.
	Zhao et al. \cite{zhao2023facial} highlighted key research directions that include MiE AU detection and MiE generation, and explored practical applications (e.g., covert emotion recognition and professional training). 
	Additionally, Zhao et al. \cite{zhao2023facial} identified pressing challenges in wild MiE applications, such as, data privacy, data protection, fairness, diversity, and the regulated use of MiE technologies.

	Previous surveys on MiE analysis have comprehensively examined the field from various perspectives, e.g., MiE analysis, dataset creation, and psychological foundations of MiE development. 
	However, critical aspects such as advanced MiE analysis and practical MiE applications remain underexplored. To address the aforementioned gaps, our work delves into state-of-the-art MiE analysis in Section \ref{commier}. Additionally, we explore the transformative potential of MiE applications in the IoT system in Section \ref{mieiot}. 
	% We expect to provide a holistic perspective that can bridge foundational research with cutting-edge advancements and wild applicability.

Based on the previous discussion, we can summarize our unique contributions over the prior research on both MaE and MiE surveys as follows: 1). While existing surveys treat MaE and MiE in isolation, our work bridges this gap by providing a \textit{holistic framework} that integrates both, addressing their complementary roles in facial expression analysis; 2). We propose a learning paradigm-based framework that unifies emerging techniques in deep MaE recognition and holographic MiE analysis that enable the systematic and insightful comparisons across these fields; 3). While much of the existing research emphasizes theoretical advancements in facial expression analysis, our work distinguishes itself by focusing on its practical deployment in \textit{real-world IoT scenarios}, with a focus on potential applications and key implementation challenges.
By unifying fundamental research, emerging technologies, and practical applications, our work offers a holistic framework that advances facial expression analysis within IoT systems. We envision that our work can foster continuous innovation by enhancing the development of emotion-aware applications across domains, e.g., smart healthcare and security.

\begin{table}[ht]\scriptsize
	\caption{An Overview of MaE Datasets}\label{tb:Mac_Datasets}
	\vspace{-0.4 cm}
	\resizebox{\textwidth}{!}{
		\renewcommand{\arraystretch}{1.1}
		\begin{tabular}{|cl|cc|ccl|cccc|ccl|}
			\hline
			\multicolumn{2}{|c|}{\multirow{2}{*}{Dataset}} & \multicolumn{2}{c|}{Subjects}               & \multicolumn{3}{c|}{MaE samples}                                                                                                                                         & \multicolumn{4}{c|}{Annotation}                                                                                                                                            & \multicolumn{3}{c|}{Collection Details}                                                        \\ \cline{3-14} 
			\multicolumn{2}{|c|}{}                         & \multicolumn{1}{c|}{Num}      & Ethnicities & \multicolumn{1}{c|}{Num}                                                                      & \multicolumn{2}{c|}{\begin{tabular}[c]{@{}c@{}}Resolution/\\ Frame rate\end{tabular}} & \multicolumn{1}{c|}{Emotion classes}                                                                        & \multicolumn{1}{c|}{FACS} & \multicolumn{1}{c|}{AU}  & Index & \multicolumn{1}{c|}{Condittion} & \multicolumn{1}{c|}{Elicitation} & Year                      \\ \hline
			\multicolumn{2}{|c|}{JAFFE \cite{670949}}                    & \multicolumn{1}{c|}{10}       & 1           & \multicolumn{1}{c|}{213 images}                                                               & \multicolumn{2}{c|}{256 × 256}                                                        & \multicolumn{1}{c|}{BEs}                                                                                    & \multicolumn{1}{c|}{N/A}  & \multicolumn{1}{c|}{N/A} & N/A   & \multicolumn{1}{c|}{Lab}        & \multicolumn{1}{c|}{P}           & 1998                      \\ \hline
			\multicolumn{2}{|c|}{KDEF \cite{goeleven2008karolinska}}                     & \multicolumn{1}{c|}{70}       & N/A         & \multicolumn{1}{c|}{4900 images}                                                              & \multicolumn{2}{c|}{562 × 762}                                                        & \multicolumn{1}{c|}{BEs}                                                                                    & \multicolumn{1}{c|}{N/A}  & \multicolumn{1}{c|}{N/A} & NA    & \multicolumn{1}{c|}{Lab}        & \multicolumn{1}{c|}{P \& S}      & 1998                      \\ \hline
			\multicolumn{2}{|c|}{MMI \cite{valstar2010induced,1521424}}                      & \multicolumn{1}{c|}{75}       & 3           & \multicolumn{1}{c|}{\begin{tabular}[c]{@{}c@{}}704 images \&\\  2900 sequences\end{tabular}}     & \multicolumn{2}{c|}{720 × 576}                                                        & \multicolumn{1}{c|}{BEs}                                                                                    & \multicolumn{1}{c|}{Yes}  & \multicolumn{1}{c|}{Yes} & NAF   & \multicolumn{1}{c|}{Lab}        & \multicolumn{1}{c|}{P}           & 2005                      \\ \hline
			\multicolumn{2}{|c|}{BU-3DFE \cite{1613022}}                  & \multicolumn{1}{c|}{100}      & 6           & \multicolumn{1}{c|}{\begin{tabular}[c]{@{}c@{}}2500 3D\\ images\end{tabular}}                 & \multicolumn{2}{c|}{1040 × 1329}                                                      & \multicolumn{1}{c|}{BEs}                                                                                    & \multicolumn{1}{c|}{N/A}  & \multicolumn{1}{c|}{N/A} & N/A   & \multicolumn{1}{c|}{Lab}        & \multicolumn{1}{c|}{P}           & 2006                      \\ \hline
			\multicolumn{2}{|c|}{TFD \cite{susskind2010toronto}}                      & \multicolumn{1}{c|}{N/A}      & N/A         & \multicolumn{1}{c|}{112,234 images}                                                           & \multicolumn{2}{c|}{48 × 48}                                                          & \multicolumn{1}{c|}{BEs}                                                                                    & \multicolumn{1}{c|}{N/A}  & \multicolumn{1}{c|}{N/A} & N/A   & \multicolumn{1}{c|}{Lab}        & \multicolumn{1}{c|}{P}           & 2010                      \\ \hline
			\multicolumn{2}{|c|}{CK+ \cite{lucey2010extended}}                      & \multicolumn{1}{c|}{123}      & 3           & \multicolumn{1}{c|}{593 sequences}                                                            & \multicolumn{2}{c|}{640 × 490}                                                        & \multicolumn{1}{c|}{BEs}                                                                                    & \multicolumn{1}{c|}{Yes}  & \multicolumn{1}{c|}{Yes} & NA    & \multicolumn{1}{c|}{Lab}        & \multicolumn{1}{c|}{P \& S}      & 2010                      \\ \hline
			\multicolumn{2}{|c|}{RaFD \cite{langner2010presentation}}                     & \multicolumn{1}{c|}{67}       & 1           & \multicolumn{1}{c|}{8040 images}                                                              & \multicolumn{2}{c|}{256 × 256}                                                        & \multicolumn{1}{c|}{\begin{tabular}[c]{@{}c@{}}BEs plus contempt \\ without neutral\end{tabular}}                                                                             & \multicolumn{1}{c|}{N/A}  & \multicolumn{1}{c|}{Yes} & N/A   & \multicolumn{1}{c|}{Lab}        & \multicolumn{1}{c|}{P}           & 2010                      \\ \hline
			\multicolumn{2}{|c|}{MUG \cite{5617662}}                      & \multicolumn{1}{c|}{86}       & 1           & \multicolumn{1}{c|}{1462 sequences}                                                           & \multicolumn{2}{c|}{896 × 896, 16fps}                                                 & \multicolumn{1}{c|}{BEs}                                                                                    & \multicolumn{1}{c|}{N/A}  & \multicolumn{1}{c|}{N/A} & NAF   & \multicolumn{1}{c|}{Lab}        & \multicolumn{1}{c|}{P}           & 2010                      \\ \hline
			\multicolumn{2}{|c|}{Multi-PIE \cite{GROSS2010807}}                & \multicolumn{1}{c|}{337}      & 4           & \multicolumn{1}{c|}{755,370 images}                                                           & \multicolumn{2}{c|}{3072 × 2048}                                                      & \multicolumn{1}{c|}{\begin{tabular}[c]{@{}c@{}}smile, surprise, squint,\\ disgust, scream and neutral\end{tabular}}         & \multicolumn{1}{c|}{N/A}  & \multicolumn{1}{c|}{N/A} & N/A   & \multicolumn{1}{c|}{Lab}        & \multicolumn{1}{c|}{P}           & 2010                      \\ \hline
			\multicolumn{2}{|c|}{SFEW 2.0 \cite{6130508}}                 & \multicolumn{1}{c|}{N/A}      & N/A         & \multicolumn{1}{c|}{1766 images}                                                              & \multicolumn{2}{c|}{Multiple}                                                         & \multicolumn{1}{c|}{BES}                                                                                    & \multicolumn{1}{c|}{N/A}  & \multicolumn{1}{c|}{N/A} & N/A   & \multicolumn{1}{c|}{Movie}      & \multicolumn{1}{c|}{P \& S}      & 2011                      \\ \hline
			\multicolumn{2}{|c|}{Oulu-CASIA \cite{ZHAO2011607}}               & \multicolumn{1}{c|}{80}       & 2           & \multicolumn{1}{c|}{\begin{tabular}[c]{@{}c@{}}1920 images \&\\  2880 sequences\end{tabular}} & \multicolumn{2}{c|}{320 × 240}                                                        & \multicolumn{1}{c|}{BEs without neutral}                                                                         & \multicolumn{1}{c|}{N/A}  & \multicolumn{1}{c|}{N/A} & NA   & \multicolumn{1}{c|}{Lab}        & \multicolumn{1}{c|}{P}           & 2011                      \\ \hline
			\multicolumn{2}{|c|}{FER-2013 \cite{goodfellow2013challenges}}                  & \multicolumn{1}{c|}{N/A}      & N/A         & \multicolumn{1}{c|}{35762 imgaes}                                                             & \multicolumn{2}{c|}{48 × 48}                                                          & \multicolumn{1}{c|}{BEs}                                                                                    & \multicolumn{1}{c|}{N/A}  & \multicolumn{1}{c|}{N/A} & N/A   & \multicolumn{1}{c|}{Internet}       & \multicolumn{1}{c|}{P \& S}      & 2012                      \\ \hline
			\multicolumn{2}{|c|}{ISED \cite{7320978}}                     & \multicolumn{1}{c|}{50}       & 1           & \multicolumn{1}{c|}{428 sequences}                                                            & \multicolumn{2}{c|}{1920 × 1080, 50fps}                                               & \multicolumn{1}{c|}{\begin{tabular}[c]{@{}c@{}}sadness, surprise, happiness\\ and disgust\end{tabular}}                                          & \multicolumn{1}{c|}{N/A}  & \multicolumn{1}{c|}{N/A} & N/A   & \multicolumn{1}{c|}{Lab}        & \multicolumn{1}{c|}{S}           & 2015                      \\ \hline
			\multicolumn{2}{|c|}{EmotioNet \cite{Benitez-Quiroz_2016_CVPR}}                & \multicolumn{1}{c|}{N/A}      & N/A         & \multicolumn{1}{c|}{950,000 images}                                                         & \multicolumn{2}{c|}{Multiple}                                                         & \multicolumn{1}{c|}{\begin{tabular}[c]{@{}c@{}}23 basic expressions or\\ compound expressions\end{tabular}} & \multicolumn{1}{c|}{N/A}  & \multicolumn{1}{c|}{Yes} & N/A   & \multicolumn{1}{c|}{Internet}   & \multicolumn{1}{c|}{P \& S}      & 2016                      \\ \hline
			\multicolumn{2}{|c|}{RAF-DB \cite{li2017reliable}}                   & \multicolumn{1}{c|}{N/A}      & N/A         & \multicolumn{1}{c|}{29,672 images}                                                            & \multicolumn{2}{c|}{Multiple}                                                        & \multicolumn{1}{c|}{\begin{tabular}[c]{@{}c@{}}BEs and twelve\\ compound expressions\end{tabular}}          & \multicolumn{1}{c|}{N/A}  & \multicolumn{1}{c|}{N/A} & N/A   & \multicolumn{1}{c|}{Internet}   & \multicolumn{1}{c|}{P \& S}      & \multicolumn{1}{c|}{2016} \\ \hline
			\multicolumn{2}{|c|}{AffectNet \cite{8013713}}                & \multicolumn{1}{c|}{N/A}      & N/A         & \multicolumn{1}{c|}{450,000 images}                                                           & \multicolumn{2}{c|}{Multiple}                                                         & \multicolumn{1}{c|}{\begin{tabular}[c]{@{}c@{}}BEs, arousal \\ and valence\end{tabular}}                    & \multicolumn{1}{c|}{N/A}  & \multicolumn{1}{c|}{N/A} & N/A   & \multicolumn{1}{c|}{Internet}   & \multicolumn{1}{c|}{P \& S}      & \multicolumn{1}{c|}{2017} \\ \hline
			\multicolumn{2}{|c|}{AFEW \cite{10.1145/3136755.3143004}}                 & \multicolumn{1}{c|}{N/A}      & N/A         & \multicolumn{1}{c|}{1809 sequences}                                                              & \multicolumn{2}{c|}{Multiple}                                                         & \multicolumn{1}{c|}{BEs}                                                                                    & \multicolumn{1}{c|}{N/A}  & \multicolumn{1}{c|}{N/A} & N/A   & \multicolumn{1}{c|}{Movie}      & \multicolumn{1}{c|}{P \& S}      & \multicolumn{1}{c|}{2017} \\ \hline
			\multicolumn{2}{|c|}{ExpW \cite{zhang2018facial}}                     & \multicolumn{1}{c|}{N/A}      & N/A         & \multicolumn{1}{c|}{91,793 images}                                                            & \multicolumn{2}{c|}{Multiple}                                                         & \multicolumn{1}{c|}{BEs}                                                                                    & \multicolumn{1}{c|}{N/A}  & \multicolumn{1}{c|}{N/A} & N/A   & \multicolumn{1}{c|}{Internet}   & \multicolumn{1}{c|}{P \& S}      & \multicolumn{1}{c|}{2018} \\ \hline
			\multicolumn{2}{|c|}{Aff-Wild2 \cite{kollias2019expression}}                & \multicolumn{1}{c|}{554}      & N/A         & \multicolumn{1}{c|}{564 sequences}                                                               & \multicolumn{2}{c|}{Multiple}                                                         & \multicolumn{1}{c|}{\begin{tabular}[c]{@{}c@{}}BEs,  arousal \\ and valence\end{tabular}}                   & \multicolumn{1}{c|}{N/A}  & \multicolumn{1}{c|}{Yes} & N/A   & \multicolumn{1}{c|}{Internet}   & \multicolumn{1}{c|}{P \& S}      & \multicolumn{1}{c|}{2019} \\ \hline
			\multicolumn{2}{|c|}{CAER \cite{Lee_2019_ICCV}}                     & \multicolumn{1}{c|}{N/A}      & N/A         & \multicolumn{1}{c|}{13,201 sequences}                                                         & \multicolumn{2}{c|}{Multiple}                                                         & \multicolumn{1}{c|}{BEs}                                                                                    & \multicolumn{1}{c|}{N/A}  & \multicolumn{1}{c|}{N/A} & N/A   & \multicolumn{1}{c|}{TV show}    & \multicolumn{1}{c|}{P \& S}      & \multicolumn{1}{c|}{2019} \\ \hline
			\multicolumn{2}{|c|}{DFEW \cite{10.1145/3394171.3413620}}                     & \multicolumn{1}{c|}{N/A}      & N/A         & \multicolumn{1}{c|}{16,372 sequences}                                                         & \multicolumn{2}{c|}{Multiple}                                                         & \multicolumn{1}{c|}{BEs}                                                                                    & \multicolumn{1}{c|}{N/A}  & \multicolumn{1}{c|}{N/A} & N/A   & \multicolumn{1}{c|}{Movie}      & \multicolumn{1}{c|}{P \& S}      & \multicolumn{1}{c|}{2020} \\ \hline
			\multicolumn{2}{|c|}{FERV39k \cite{Wang_2022_CVPR}}                  & \multicolumn{1}{c|}{N/A}      & N/A         & \multicolumn{1}{c|}{38,935 sequences}                                                         & \multicolumn{2}{c|}{3840 × 2160}                                                      & \multicolumn{1}{c|}{BEs}                                                                                    & \multicolumn{1}{c|}{N/A}  & \multicolumn{1}{c|}{N/A} & N/A   & \multicolumn{1}{c|}{Internet}   & \multicolumn{1}{c|}{P \& S}      & \multicolumn{1}{c|}{2021} \\ \hline
			\multicolumn{2}{|c|}{PEDFE \cite{miolla2023padova}}                    & \multicolumn{1}{c|}{56}       & N/A         & \multicolumn{1}{c|}{1458 sequences}                                                           & \multicolumn{2}{c|}{1920 × 1080, 30fps}                                               & \multicolumn{1}{c|}{BEs without neutral}                                                                         & \multicolumn{1}{c|}{N/A}  & \multicolumn{1}{c|}{N/A} & NA    & \multicolumn{1}{c|}{Lab}        & \multicolumn{1}{c|}{P \& S}      & \multicolumn{1}{c|}{2022} \\ \hline
			\multicolumn{2}{|c|}{CalD3r + MenD3s \cite{ULRICH2024104033}}          & \multicolumn{1}{c|}{104 + 92} & 1+1         & \multicolumn{1}{c|}{\begin{tabular}[c]{@{}c@{}}4,678 images\\ 4,038 images\end{tabular}}      & \multicolumn{2}{c|}{640 × 480, 30fps}                                                 & \multicolumn{1}{c|}{BEs}                                                                                    & \multicolumn{1}{c|}{N/A}  & \multicolumn{1}{c|}{N/A} & N/A   & \multicolumn{1}{c|}{Lab}        & \multicolumn{1}{c|}{S}           & \multicolumn{1}{c|}{2024} \\ \hline
		\end{tabular}
	}
	\begin{tablenotes}\scriptsize
		\item $^1$ BEs: Basic expressions (seven); N/A: Not applicable; P: Posed; S: Spontaneous; NAF: Onset frame, Apex frame, Offset frame, respectively.
	\end{tablenotes}
	\vspace{-0.5 cm}
\end{table}

	\vspace{-0.5 cm}
	\section{Datasets for Facial Expression Analysis}\label{dataset}
	Facial expression analysis relies on high-quality datasets to drive advancements in model development and evaluation. 
	In this section, we will provide a holistic introduction to the current datasets used for MaE and MiE analysis.
	
	\vspace{-0.5 cm}
	\subsection{MaE Datasets}\vspace{-0.1 cm}
	The recent MaE datasets including Japanese female facial expression (JAFFE) \cite{670949}, Karolinska directed emotional faces (KDEF) \cite{goeleven2008karolinska}, MMI \cite{valstar2010induced,1521424}, Binghamton university 3D facial expression (BU-3DFE) \cite{1613022}, the Toronto face database (TFD) \cite{susskind2010toronto}, The Extended Cohn-Kanade Dataset (CK+) \cite{lucey2010extended}, Radboud faces database (RaFD) \cite{langner2010presentation}, MUG \cite{5617662}, Multi-PIE \cite{GROSS2010807}, static facial expressions in the wild (SFEW 2.0) \cite{6130508}, Oulu-CASIA \cite{ZHAO2011607}, facial expression recognition 2013 (FER-2013) \cite{goodfellow2013challenges}, Indian spontaneous expression database (ISED) \cite{7320978}, EmotioNet \cite{Benitez-Quiroz_2016_CVPR}, wild affective faces database (RAF-DB) \cite{li2017reliable}, AffectNet \cite{8013713}, the acted facial expressions in the wild (AFEW) \cite{10.1145/3136755.3143004}, expression in-the-wild (ExpW) \cite{zhang2018facial}, Aff-Wild2 \cite{kollias2019expression}, context-aware emotion recognition (CAER) \cite{Lee_2019_ICCV}, dynamic facial expression in the wild (DFEW) \cite{10.1145/3394171.3413620}, FERV39k \cite{Wang_2022_CVPR}, Padova emotional dataset of facial expressions (PEDFE) \cite{miolla2023padova}, CalD3r + MenD3s \cite{ULRICH2024104033}. 
	Table~\ref{tb:Mac_Datasets} illustrates the detailed information of the MaE datasets.

	The basic expressions of most MaE datasets can be divided into seven categories, i.e., anger, neutral, disgust, fear, happiness, sadness, and surprise. 
	Although being considered as a basic emotion of the face, contempt expression is not common in most MaE datasets. 
	Few datasets (e.g., CK+ and RaFD) have collected the contempt expression; therefore, the number of samples on contempt expression is limited. 
	In the CK+ dataset, the number of contempt expression samples accounts for 5\% of total samples.
	In addition to the seven basic expressions in MaE datasets, several datasets (e.g., AffectNet and AFF-wild2) also used the continuous-valued valence and arousal to describe the intensity of expression. 
	Moreover, only the ExpW dataset considered the compound expressions. 
	In the ExpW dataset, 23 basic or compound emotion categories were used for emotion classification. 
	
	The early datasets were usually collected from the laboratory and required a certain number of subjects to join. The collection of these datasets required a lot of labor, which may include subjects participating in the collection, guides of collection experiments, and professional annotators. In addition to the annotation of expression categories, several datasets (e.g., CK+, MMI, and Oulu-CASIA) required additional information in terms of facial action coding system (FACS), AU, and index. Typically, CK+ and Oulu-CASIA have utilized the annotation information of the index to label their sequences with the onset and peak of expressions. MMI and CK+ have labeled the FACS and AU to provide more additional information. In addition to the labor-consuming collection of MaE datasets, EmotioNet has utilized the automatic annotation algorithm to label the data collected from the internet. In this dataset, a total of 9500,000 images were annotated with AUs, AU intensities, and emotion categories.

	The early MaE datasets (e.g., JAFFE) were utilized to obtain the posed images and sequences. These datasets contained the posed data for the front view. To meet the requirement of practical MaE in wild conditions, many datasets (e.g., Multi-PIE, RaFD) provided the MaE data with various environmental conditions including multiple head poses, occlusions, and illuminations. Typically, the Multi-PIE contained MaE images under 19 illumination and 15 viewpoint conditions in four sessions. In their process of dataset collection, each subject was recorded between $-90^\circ$ to $90^\circ$ with an interval of $15^\circ$. MaE in RaFD was recorded at the same moment through five different camera angles and shown with three different gaze directions. In addition, the MaE datasets collected from the internet (e.g., AffectNet and ExpW) contained an amount of MaE data with more wild scenarios. These datasets can benefit the robustness and generalization of further MaE research.

	\vspace{-0.4 cm}
	\begin{table}[ht]\scriptsize
		\caption{Spontaneous MiE Datasets}\label{tb:Datasets}
		\vspace{-0.4 cm}
		\resizebox{\textwidth}{!}{
			\renewcommand{\arraystretch}{1.1}
			\begin{tabular}{|cl|ccc|ccc|cccc|}
				\hline
				\multicolumn{2}{|c|}{\multirow{2}{*}{Dataset}}    & \multicolumn{3}{c|}{Subjects}                                                             & \multicolumn{3}{c|}{MiE Videos}                                                                                           & \multicolumn{4}{c|}{Annotations}                                                                                                                                                                                                                \\ \cline{3-12} 
				\multicolumn{2}{|c|}{}                            & \multicolumn{1}{c|}{Num} & \multicolumn{1}{c|}{Mean age}             & Ethnicities        & \multicolumn{1}{c|}{Num}  & \multicolumn{1}{c|}{Resolution}                                                               & Frame rate & \multicolumn{1}{c|}{Emotion classes}                                                                                                             & \multicolumn{1}{c|}{FACS}                & \multicolumn{1}{c|}{AU}                   & Index \\ \hline
				\multicolumn{1}{|c|}{\multirow{3}{*}{SMIC \cite{li2013spontaneous}}} & HS  & \multicolumn{1}{c|}{16}  & \multicolumn{1}{c|}{\multirow{3}{*}{N/A}} & \multirow{3}{*}{3} & \multicolumn{1}{c|}{164}  & \multicolumn{1}{c|}{\multirow{3}{*}{640$\times$480}}                                                  & 100        & \multicolumn{1}{c|}{Pos (51) Neg (70) Sur (43)}                                                                                                  & \multicolumn{1}{c|}{\multirow{3}{*}{N/A}} & \multicolumn{1}{c|}{\multirow{3}{*}{N/A}} & NF    \\ \cline{2-3} \cline{6-6} \cline{8-9} \cline{12-12} 
				\multicolumn{1}{|c|}{}                      & VIS & \multicolumn{1}{c|}{8}   & \multicolumn{1}{c|}{}                     &                    & \multicolumn{1}{c|}{71}   & \multicolumn{1}{c|}{}                                                                         & 25         & \multicolumn{1}{c|}{Pos (28) Neg (23) Sur (20)}                                                                                                  & \multicolumn{1}{c|}{}                    & \multicolumn{1}{c|}{}                     & N/A   \\ \cline{2-3} \cline{6-6} \cline{8-9} \cline{12-12} 
				\multicolumn{1}{|c|}{}                      & NIR & \multicolumn{1}{c|}{8}   & \multicolumn{1}{c|}{}                     &                    & \multicolumn{1}{c|}{71}   & \multicolumn{1}{c|}{}                                                                         & 25         & \multicolumn{1}{c|}{Pos (28) Neg (24) Sur (19)}                                                                                                  & \multicolumn{1}{c|}{}                    & \multicolumn{1}{c|}{}                     & N/A   \\ \hline
				\multicolumn{2}{|c|}{CASME \cite{yan2013casme}}                       & \multicolumn{1}{c|}{35}  & \multicolumn{1}{c|}{22.03}                & 1                  & \multicolumn{1}{c|}{195}  & \multicolumn{1}{c|}{\begin{tabular}[c]{@{}c@{}}640$\times$480\\ 1280$\times$720\end{tabular}} & 60         & \multicolumn{1}{c|}{\begin{tabular}[c]{@{}c@{}}Hap (5) Dis (88) Sad (6) Con (3)\\ Fea (2) Anx (28) Sur (20) Rep (40)\end{tabular}}               & \multicolumn{1}{c|}{Yes}                 & \multicolumn{1}{c|}{Yes}                  & NAF   \\ \hline
				\multicolumn{2}{|c|}{CASME II \cite{yan2014casme}}                    & \multicolumn{1}{c|}{35}  & \multicolumn{1}{c|}{22.03}                & 1                  & \multicolumn{1}{c|}{247}  & \multicolumn{1}{c|}{640$\times$480}                                              & 200        & \multicolumn{1}{c|}{\begin{tabular}[c]{@{}c@{}}Hap (33) Sur (25) Dis (60) \\ Rep (27) Oth (102)\end{tabular}}                                    & \multicolumn{1}{c|}{Yes}                 & \multicolumn{1}{c|}{Yes}                  & NAF   \\ \hline
				\multicolumn{2}{|c|}{CAS(ME)\textsuperscript{2} \cite{qu2017cas}}                     & \multicolumn{1}{c|}{22}  & \multicolumn{1}{c|}{22.59}                & 1                  & \multicolumn{1}{c|}{57}   & \multicolumn{1}{c|}{640$\times$480}                                              & 30         & \multicolumn{1}{c|}{Hap (51) Neg (70) Sur (43) Oth (19)}                                                                                         & \multicolumn{1}{c|}{Yes}                 & \multicolumn{1}{c|}{Yes}                  & NAF   \\ \hline
				\multicolumn{2}{|c|}{SAMM \cite{davison2016samm}}                        & \multicolumn{1}{c|}{32}  & \multicolumn{1}{c|}{33.24}                & 13                 & \multicolumn{1}{c|}{159}  & \multicolumn{1}{c|}{2040$\times$1088}                                            & 200        & \multicolumn{1}{c|}{\begin{tabular}[c]{@{}c@{}}Hap (24) Dis (8) Sad (3) Fea (7) \\ Sur (13) Ang (20) Oth (84)\end{tabular}}                      & \multicolumn{1}{c|}{Yes}                 & \multicolumn{1}{c|}{Yes}                  & NAF   \\ \hline
				\multicolumn{2}{|c|}{MEVIEW \cite{husak2017spotting}}                      & \multicolumn{1}{c|}{16}  & \multicolumn{1}{c|}{N/A}                  & N/A                & \multicolumn{1}{c|}{40}   & \multicolumn{1}{c|}{1280$\times$720}                                             & 25         & \multicolumn{1}{c|}{\begin{tabular}[c]{@{}c@{}}Hap (6) Dis (1) Fea (3) Sur (9) \\ Con (6) Ang (2) Conf (13)\end{tabular}}                         & \multicolumn{1}{c|}{Yes}                 & \multicolumn{1}{c|}{Yes}                  & NF    \\ \hline
				\multicolumn{2}{|c|}{CAS(ME)\textsuperscript{3} \cite{li2022cas}}                     & \multicolumn{1}{c|}{247} & \multicolumn{1}{c|}{22.74}                & 1                  & \multicolumn{1}{c|}{1059} & \multicolumn{1}{c|}{1280$\times$720}                                             & 30         & \multicolumn{1}{c|}{\begin{tabular}[c]{@{}c@{}}Hap (992) Dis (2528) Sad (635) Fea (892) \\ Sur (1208) Con (401) Ang (20) Oth (251)\end{tabular}} & \multicolumn{1}{c|}{Yes}                 & \multicolumn{1}{c|}{Yes}                  & NAF   \\ \hline
				\multicolumn{2}{|c|}{MMEW \cite{ben2021video}}                        & \multicolumn{1}{c|}{36}  & \multicolumn{1}{c|}{N/A}                  & 1                  & \multicolumn{1}{c|}{16}   & \multicolumn{1}{c|}{1920$\times$1080}                                            & 25         & \multicolumn{1}{c|}{\begin{tabular}[c]{@{}c@{}}Hap (36) Dis (72) Sad (13) Fea (16) \\ Sur (80) Ang (8) Oth (84)\end{tabular}}                    & \multicolumn{1}{c|}{Yes}                 & \multicolumn{1}{c|}{Yes}                  & NAF   \\ \hline
				\multicolumn{2}{|c|}{4DME \cite{li20224dme}}                        & \multicolumn{1}{c|}{56}  & \multicolumn{1}{c|}{27.8}                 & 3                  & \multicolumn{1}{c|}{1068} & \multicolumn{1}{c|}{\begin{tabular}[c]{@{}c@{}}640$\times$480\\ 1200$\times$1600\end{tabular}}                                             & {\begin{tabular}[c]{@{}c@{}}30/60\\ 60\end{tabular}}         & \multicolumn{1}{c|}{\begin{tabular}[c]{@{}c@{}}Pos (34) Neg (127) Sur(30) Rep(6)\\ PS(13) NS(8) RS (3) PR(8) NR (7) Oth(31)\end{tabular}}        & \multicolumn{1}{c|}{Yes}                 & \multicolumn{1}{c|}{Yes}                  & NAF   \\ \hline
		\end{tabular}}
		\begin{tablenotes}\footnotesize
			\item $^1$Pos: Positivity; Neg: negativity; Sur: Surprise; Hap: Happiness; Dis: Disgust; Sad: Sadness; Con: Contempt; Fea: Fear; Anx: Anxiety; Rep: Repression; Ang: Anger; Conf: Confusion; Oth: Others; PS: Pos w/ Sur; NS: Neg w/ Sur; RS: Rep w/ Sur; PR: Pos w/ Rep; NR: Neg w/ Rep;     
			\item $^2$N/A: Not applicable; 	NAF: Onset frame, Apex frame, and Offset frame, respectively.
		\end{tablenotes}
		\vspace{-0.4 cm}
	\end{table}

\vspace{-0.2 cm}
    \subsection{MiE Datasets}\vspace{-0.1 cm}
	In recent years, multiple datasets have been built for further development of MiE analysis. Initially, researchers collected the MiE databases (e.g., Polikvsky’s database \cite{polikovsky2009facial} and USF-HD \cite{shreve2011macro}) by asking the participants to pose or mimic facial expressions. These posed datasets supported the preliminary studies of MiE, but were not used for current analysis due to the properties of private datasets and not capturing the nature of MiE. Later, more spontaneous datasets, including the spontaneous micro-expression corpus (SMIC) \cite{li2013spontaneous}, Chinese Academy of Sciences micro-expression (CASME) \cite{yan2013casme}, Chinese Academy of Sciences micro-expression II (CASME II) \cite{yan2014casme}, Chinese Academy of Sciences macro-expressions and micro-expressions (CAS(ME)\textsuperscript{2}) \cite{qu2017cas}, spontaneous actions and micro-movements (SAMM) \cite{davison2016samm}, micro-expression videos in the wild (MEVIEW) \cite{husak2017spotting}, CAS(ME)\textsuperscript{3} \cite{li2022cas}, MMEW \cite{ben2021video}, and 4D spontaneous micro expression database (4DME) \cite{li20224dme}, were utilized in MiE research. The information on the spontaneous MiE datasets is detailed in Table \ref{tb:Datasets}.

	Due to the subtle and fleeting nature of MiE, the annotation of MiE datasets is a challenging and demanding task that requires a significant amount of time and labor. Coders who label MiE need FACS training and half an hour to detect suitable clips \cite{pantic2009machine}. The labeling of MiE includes the AU and emotion classification. By incorporating the AU into certain emotion taxonomy, the technique of MiE synthesis can be used to generate new samples. The classification of MiE in the dataset is different, which results in inconsistent annotations between various datasets. For example, the classification of MiE in SMIC is positive, negative, and surprising. The classification of MiE in other datasets, such as CASME, MEVIEW, and SAMM, can be divided into more than six categories.

	To induce spontaneous MiE, many studies have used a collection paradigm that requires participants to maintain a poker face while watching intensely emotional video clips. MiEs are revealed and recorded using a high-speed camera. 
	This effective and simplified method was applied to five datasets: SMIC, CASME, CASME II, SAMM, and CAS(ME)\textsuperscript{2}. 
	However, this paradigm was criticized for its laboratory setting, which lacks practical or realistic situations. Consequently, studies expanded to more realistic settings. 
	In real-life scenarios, such as high-stakes poker games or TV interviews, participants often conceal their true emotions, leading to MiE occurrences. 
	For example, MEVIEW constructed an in-the-wild MiE dataset from real poker games and TV interviews. 
	Despite its benefits for real-scene analysis, MEVIEW has limited data samples and frequent face pose changes, resulting in side views and occlusions. 
	Subsequently, CAS(ME)\textsuperscript{3} adopted a trade-off method using mock crime paradigms to collect MiE samples in practical scenes with controllable factors.
	
	\begin{figure}[htb]
		\vspace{-0.2 cm}
		\centerline{\includegraphics[width=0.95\linewidth]{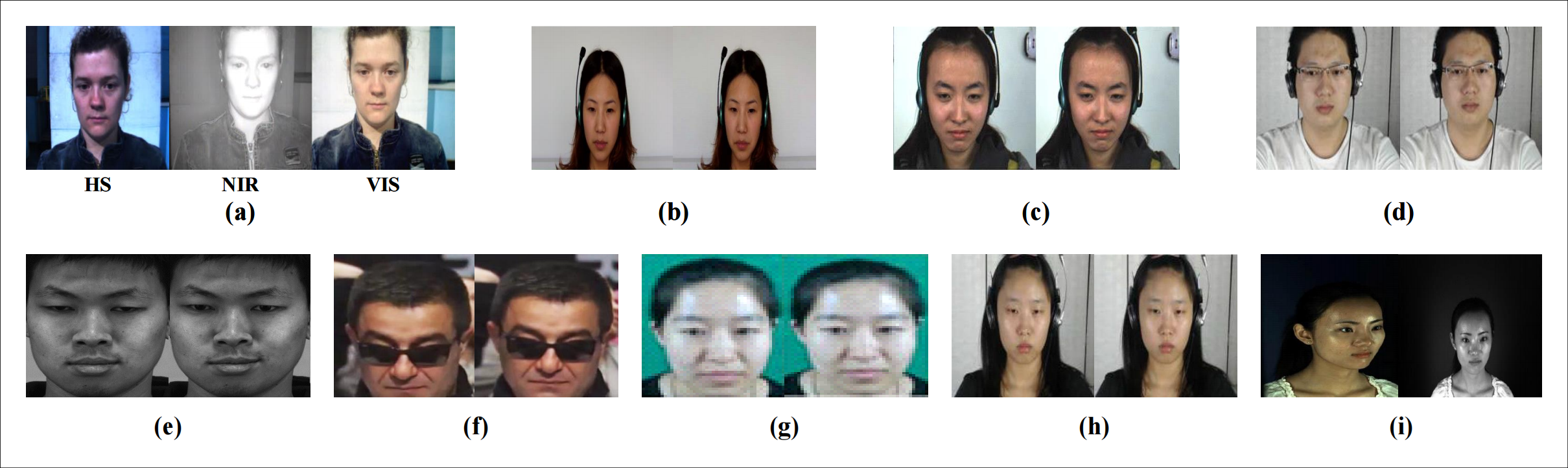}}
		\vspace{-0.2 cm}
		\caption{Samples of current MiE datasets. These are as follows: (a) SMIC \cite{li2013spontaneous}, (b) CASME \cite{yan2013casme}, (c) CASME II \cite{yan2014casme}, (d) CAS(ME)\textsuperscript{2} \cite{qu2017cas}, (e) SAMM \cite{davison2016samm}, (f) MEVIEW \cite{husak2017spotting}, (g) CAS(ME)\textsuperscript{3} \cite{li2022cas}, (h) MMEW \cite{ben2021video}, and (i) 4DME \cite{li20224dme}.}
		\Description{This figure provides the samples of current MiE datasets.}\label{fig3}
		\vspace{-0.5 cm}
	\end{figure}

	Early spontaneous MiE datasets collected the 2D facial video, which had the advantages of convenient collection and control. 
	With the concern about adequate and accurate MiE analysis, more modalities of the dataset were emphasized. 
	Except for using 2D facial video, the dataset of SMIC implemented the 2D near infrared videos to alleviate the influence of illumination. 
	More recently, the depth information has proven its effectiveness from face recognition \cite{mian2007efficient} to expression recognition \cite{corneanu2016survey}. 
	A standard RGB image can be constructed into a 3D model of the human face with depth information, which can enhance the robust perception of expression features and cognitive behavior. 
	Therefore, the modal of depth information was applied in the datasets of CAS(ME)\textsuperscript{3} and 4DME. Besides, CAS(ME)\textsuperscript{3} enriched the data information with physiological signals and voice signals. 
	Also, 4DME developed more facial multi-modal videos, consisting of reconstructed dynamic 3D facial meshes, grayscale 2D frontal facial videos, Kinect-color videos, and Kinect-depth videos. Fig.~\ref{fig3} has shown the samples in spontaneous MiE datasets.

	In addition, some datasets considered the composition of MiE and other expressions for the real-scene situation analysis. For example, MMEW, CAS(ME)\textsuperscript{2}, CAS(ME)\textsuperscript{3}, and 4DME contained the MiE and MaE, which can be employed for detecting MiE from complete videos and further analysis about the evolution of different expressions. Due to the short length of video data, the analysis of MiE was constrained. To collect long videos, several datasets, including CASME, CASME II, and SMIC, were expanded to incorporate frames that do not exhibit MiE before and after the annotated MiE samples. Long MiE videos can benefit more sharper MiE algorithms in more complex situations, such as head movement and verbal behavior that occurred in common scenarios.

\subsection{Challenges in Practical Applications of Facial Expression Datasets}
{\color{black}
	Despite the significant progress in developing MaE and MiE datasets, several challenges persist when applying them to real-world scenarios. Four key challenges that hinder the practical application of facial expression datasets are discussed as follows. \textbf{Class Imbalance and Label Inconsistency:} Overrepresentation of common expressions (e.g., happiness and sadness)  versus underrepresentation rare ones (e.g., contempt, fear, and disgust). Besides, inconsistencies in expression classification exist for both MaE and MiE datasets. 
	\textbf{Collection Complexity:} Facial expression data collection poses substantial challenges. Spontaneous expressions are inherently more difficult to elicit and capture than posed ones, making the collection of spontaneous datasets more difficult. Ensuring demographic diversity—such as including participants across ethnicities and age groups—further complicates the data collection process. Additionally, MiE datasets require high-frame-rate recordings to capture subtle and transient facial movements, imposing strict demands on both recording equipment and experimental design. \textbf{Annotation Complexity:} MiE labeling is very labor-intensive and subject to inter-annotator variability, even among trained experts, which further reduces reliability. Meanwhile, the large scale of MaE datasets demands automated annotation strategies to for data labeling. \textbf{Domain Shift:} Domain Shift arises both across datasets within the same task (e.g., MaE and MiE) and between training data and real-world deployment scenarios. Variations in collection protocols, environmental conditions, and label annotation often lead to inconsistent model performance across datasets. Moreover, the mismatch between lab-controlled data and in-the-wild expressions further hinders the robustness of real-world applications.
	}

	% \vspace{-0.4 cm}
	\begin{table}[htb]\scriptsize
		\caption{An Overview of Representative Methods on Deep Static MaE Recognition}\label{tb:MaERST}
		\vspace{-0.3 cm}
		\resizebox{\textwidth}{!}{
			\renewcommand{\arraystretch}{1.1}
			\begin{tabular}{|c|l|c|c|c|c|c|c|c|}
			\hline
			\begin{tabular}[c]{@{}c@{}}Learning \\ Paradigms\end{tabular} & \multicolumn{1}{c|}{Method} & Year & Pre-processing & Block           & Dataset                                                                                    & Performance                                                                        & \begin{tabular}[c]{@{}c@{}}Network\\ Structure\end{tabular} & Protocol           \\ \hline
			\multirow{11}{*}{Ensemble Learning}           & Kim et al. \cite{10.1145/2818346.2830590}                 & 2016 & IN+FA+CE       & N/A             & SFEW 2.0                                                                                   & 61.6\%                                                                             & CNN                                                         & Hold out           \\ \cline{2-9} 
			& Pons et.al . \cite{8039231}               & 2018 & IN+FA          & N/A             & SFEW 2.0                                                                                   & 42.9\%                                                                             & CNN                                                         & LOSO               \\ \cline{2-9} 
			& Saurav et al. \cite{saurav2022dual}              & 2022 & IN+FA+CE       & N/A             & FER2013/RAF-DB/CK+                                                              & 72.77\%/86.07\%/98.54\%                                                  & CNN                                                         & LOSO               \\ \cline{2-9} 
			& Shao et al. \cite{SHAO201982}               & 2019 & FD+FA+DA       & RES             & CK+/BU-3DFE/FER2013                                                                        & 95.29\%/86.50\%/71.14\%                                                            & CNN                                                         & 10 fold            \\ \cline{2-9} 
			& Reddy et al. \cite{VISWANATHAREDDY202023}             & 2020 & DA             & N/A             & AffectNet                                                                                  & 59\%                                                                               & CNN                                                         & Hold out           \\ \cline{2-9} 
			& Sharifnejad et al. \cite{asd11}         & 2021 & FD             & N/A             & CK+                                                                                        & 95.33\%                                                                            & N/A                                                         & 10 fold            \\ \cline{2-9} 
			& Hariri et al. \cite{HARIRI202184}              & 2021 & N/A            & N/A             & BU-3DFE                                                                                    & 94.73\%                                                                            & CNN                                                         & 10 fold            \\ \cline{2-9} 
			& Mohan et al. \cite{9226437}               & 2021 & DA             & N/A             & CK+/JAFFE/KDEF/FER2013/RAFDB                                                               & 98\%/98\%/96\%/78\%/83\%                                                           & 2sCNN                                                       & 10 fold            \\ \cline{2-9} 
			& Xie et al. \cite{8371638}                 & 2019 & FD             & N/A             & CK+/JAFFE                                                                                  & 93.46\%/94.75\%                                                                    & 2sCNN                                                       & 10 fold            \\ \cline{2-9} 
			& Wadhawan et al. \cite{9767587}            & 2023 & FD+CE          & N/A             & CK+/JAFFE                                                                            & 97.31\%/97.14\%                                                            & CNN                                                         & 10 fold+LOSO       \\ \cline{2-9} 
			& Hua et al. \cite{8643924}                 & 2019 & DA             & N/A             & FER2013/JAFFE/AffectNet                                                                    & 71.91\%/96.44\%/62.11\%                                                            & 3sCNN                                                       & 10 fold            \\ \hline
			\multirow{4}{*}{Transfer Learning}                            & Ng et al. \cite{10.1145/2818346.2830593}                  & 2015 & FD             & N/A             & Emotion Recognition In The Wild Challenge                                                                                     & 55.6\%                                                                             & CNN                                                         & 10 fold            \\ \cline{2-9} 
			& Akhand et al. \cite{electronics10091036}              & 2021 & FD+DA          & N/A             & KDEF/JAFFE                                                                                 & 96.51\%/99.52\%                                                                    & CNN                                                         & Hold out           \\ \cline{2-9} 
			& Ngo et al. \cite{s20092639}                 & 2020 & DA             & RES             & AffectNet                                                                                  & 60.70\%                                                                            & CNN                                                         & Hold out           \\ \cline{2-9} 
			& Atabansi et al. \cite{Atabansi_2021}            & 2021 & FD+DA          & N/A             & Oulu-CASIA NIR                                                                                 & 98.11\%                                                                            & CNN                                                         & Hold out           \\ \hline
			\multirow{11}{*}{Multi-task Learning}                         & Pons et al. \cite{pons2018multi}                & 2018 & DA             & RES             & SFEW 2.0                                                                                   & 45.9\%                                                                             & CNN                                                         & Hold out           \\ \cline{2-9} 
			& Chen et al. \cite{CHEN2021107893}                & 2021 & N/A            & RES             & AffectNet/RAF                                                                              & 61.98\%/87.27\%                                                                    & CNN                                                         & 10 fold            \\ \cline{2-9} 
			& Yu et al. \cite{YU2022108401}                  & 2022 & DA+FA          & RES+Atten       & RAF-DB/SFEW 2.0/CK+/Oulu-CASIA                                                             & 90.36\%/45.78\%/98.33\%/87.32\%                                                     & CNN                                                         & 10 fold+LOSO       \\ \cline{2-9} 
			& Liu et al. \cite{10214365}                 & 2023 & N/A            & RES+Trans       & AffectNet/Aff-Wild2                                                                        & 65.80\%/64.13\%                                                                      & CNN                                                         & Hold out           \\ \cline{2-9} 
			& Yu et al. \cite{yu2020facial}                  & 2020 & FA+ROI         & N/A             & CK+/Oulu-CASIA                                                                             & 99.08\%/90.40\%                                                                    & GRU                                                         & 10 fold            \\ \cline{2-9} 
			& Xiao et al.  \cite{XIAO2023110312}                  & 2023 & FD             & N/A             & CK+/MMI/RAF-DB                                                                             & 99.07\%/84.62\%/87.52\%                                                             & CNN                                                         & 5 fold+Hold out    \\ \cline{2-9} 
			& Zhao et al. \cite{9413000}                & 2021 & FA+DA          & N/A             & CK+/Oulu-CASIA/MMI                                                                         & 97.85\%/89.23\%/75.32\%                                                            & CNN                                                         & 10 fold+Hold out   \\ \cline{2-9} 
			& Chen et al. \cite{chen2022orthogonal}                & 2022 & FD+FA          & Atten           & Multi-PIE/KDEF                                                                             & 88.41\%/89.04\%                                                                    & CNN                                                         & 5 fold             \\ \cline{2-9} 
			& Pan et al. \cite{9945168}                 & 2022 & DA             & RES             & RAF-DB/FER2013                                                                             & 88.30\%/68.54\%                                                                    & CNN                                                         & Hold out           \\ \cline{2-9} 
			& Qin et al. \cite{10216308}                  & 2023 & DA             & Trans +Atten          & RAF-DB                                                                                     & 90.97\%                                                                            & CNN                                                         & Hold out           \\ \cline{2-9} 
			& Foggia et al. \cite{FOGGIA2023105651}              & 2023 & DA             & RES             & RAF-DB                                                                                     & 85.30\%                                                                   & CNN                                                         & 5 fold             \\ \hline
			\multirow{12}{*}{Attention-Based Learning}                    & Fernandez et al. \cite{Fernandez_2019_CVPR_Workshops}           & 2019 & DA             & RES+Atten       & CK+/BU-3DFE                                                                                & 90.30\%/82.11\%                                                                    & CNN                                                         & Cross-validation   \\ \cline{2-9} 
			& Li et al. \cite{LI2020340}                  & 2020 & FD             & RES+Atten       & CK+/JAFFE/Oulu-CASIA/FER2013                                                               & 98.68\%/98.52\%/94.63\%/75.82\%                                                    & 2sCNN                                                       & 5 fold             \\ \cline{2-9} 
			& Li et al. \cite{8576656}                  & 2018 & FD+FA             & RES+Atten       & RAF-DB/AffectNet                                                                           & 85.07\%/58.78\%                                                                    & CNN                                                         & 10 fold            \\ \cline{2-9} 
			& Zhao et al. \cite{9474949}                & 2021 & FA+DA          & RES+Atten       & CAER/AffectNet/RAF-DB/SFEW 2.0                                                             & 88.42\%/64.53\%/88.40\%/59.40\%                                                 & CNN                                                         & Cross-validation   \\ \cline{2-9} 
			& Liu et al. \cite{9750079}                 & 2022 & DA+ROI         & RES+Atten       & \begin{tabular}[c]{@{}c@{}}RAF-DB/AffectNet/ SFEW 2.0/FER2013\\ /AffectNet-8\end{tabular} & \begin{tabular}[c]{@{}c@{}}89.25\%/64.54\%/61.17\%/74.48\%\\ /61.74\%\end{tabular} & CNN                                                         & Hold out   \\ \cline{2-9} 
			& Aouayeb et al. \cite{aouayeb2021learning}             & 2021 & DA             & RES+Trans    & CK+/JAFFE/SFEW 2.0/RAF-DB                                                             & 99.80\%/92.92\%/54.29\%/87.22\%                                                    & CNN                                                         & 10 fold + Hold out \\ \cline{2-9} 
			& Zheng et al. \cite{Zheng_2023_ICCV}               & 2023 & FA             & RES+Trans+Atten & RAF-DB/AffectNet/FERPlus                                                                   & 92.05\%/67.31\%/91.62\%                                                            & CNN                                                         & Hold out   \\ \cline{2-9} 
			& Xue et al. \cite{Xue_2021_ICCV}                 & 2021 & FA+FD          & RES+Trans+Atten & RAF-DB/AffectNet/FERPlus                                                                   & 90.91\%/66.23\%/90.83\%                                                            & CNN                                                         & Hold out           \\ \cline{2-9} 
			& Liu et al. \cite{LIU2023781}                 & 2023 & FD+FA          & RES+Trans+Atten & RAF-DB/AffectNet/FERPlus                                                                   & 88.21\%/60.68\%/88.72\%                                                            & CNN                                                         & Hold out            \\ \cline{2-9} 
			& Meng et al. \cite{10447031}                 & 2024 & N/A            & RES+Trans+Atten & RAF-DB/AffectNet/AffectNet-8                                                               & 92.36\%/67.44\%/64.26\%                                                            & 2sCNN                                                       & Hold out           \\ \cline{2-9} 
			& Sun et al. \cite{10061322}                 & 2023 & DA             & RES+Trans+Atten & RAF-DB/AffectNet/FER2013                                                                   & 89.50\%/65.66\%/74.84\%                                                            & 2sCNN                                                       & Hold out           \\ \cline{2-9} 
			& Feng et al. \cite{10018959}                & 2023 & DA             & Trans           & RAF-DB/AffectNet/FERPlus                                                                   & 90.38\%/63.33\%/90.41\%                                                            & Trans                                                       & Hold out   \\ \hline
			\multirow{7}{*}{Self-supervised Learning}                     & Roy et al. \cite{roy2021self}                 & 2021 & DA             & RES             & KDEF/DDCF                                                                                  & 94.64\%/95.26\%                                                                    & CNN                                                         & 10 fold            \\ \cline{2-9} 
			& Roy et al. \cite{10.1145/3632960}                 & 2023 & DA             & RES             & KDEF/DDCF/BU3DFE                                                                                   & 97.15\%/97.34\%/97.02\%                                                                    & CNN                                                         & 10 fold            \\ \cline{2-9} 
			& Li et al.  \cite{9511468}                  & 2021 & FD+DA          & RES+Atten       & RAF-DB/CK+/MMI/JAFFE                                                                       & 88.23\%/98.77\%/79.42\%/91.89\%                                                    & CNN                                                         & 10 fold+Hold out   \\ \cline{2-9} 
			& Wang et al. \cite{Wang_2022_ACCV}                & 2022 & DA             & RES             & VGGFace2/RAF-DB/FED-RO                                                                     & 60.20\%/85.95\%/70.00\%                                                            & CNN                                                         & 10 fold+Hold out            \\ \cline{2-9} 
			& Chen et al. \cite{CHEN2023206}                & 2023 & DA             & Trans           & FER2013/AffectNet/SFEW 2.0/RAF-DB                                                          & 74.95\%/66.04\%/63.69\%/90.98\%                                                    & Trans                                                       & Hold out           \\ \cline{2-9} 
			& Fang et al. \cite{10121455}                & 2023 & DA             & RES             & RAF-DB/FERPlus/AffectNet                                                                   & 78.17\%/65.54\%/50.33\%                                                            & CNN                                                         & Hold out   \\ \cline{2-9} 
			& An et al. \cite{an2024self}                  & 2024 & N/A       & Trans             & RAF-DB/FERPlus/AffectNet/AffectNet-8                                                       & 91.45\%/90.16\%/63.49\%/60.75\%                                                    & Trans                                                         & Hold out   \\ \hline
		\end{tabular}
		
	}
	\begin{tablenotes}\scriptsize
		\item $ ^1$DA: Data augmentation; FD: Face detection; CE: Contrast enhancement; IN: Illumination normalization; FA: Face alignment; N/A: Not applicable; Trans: Transformer; Atten: Attention; 
		nsCNN: n-stream convolutional neural network; LOSO: Leave-one-subject-out; ROI: Region of interest; RES: Residual; GRU: Gated recurrent unit. 
	\end{tablenotes}
\vspace{-0.8 cm}
\end{table}

% \vspace{-0.1 cm}
\section{Deep MaE Recognition}\label{MaErec}\vspace{-0.1 cm}
The ever-increasing data volume has motivated various data-driven methods to handle the technical challenges in computer vision, such as, emotion recognition \cite{li2020deep}, event detection \cite{hu2022event} and action detection \cite{vahdani2022deep}.
As an important research direction in computer vision, deep learning-based MaE recognition (a.k.a. deep MaE recognition) has seen its significant advancements in both practical applications and research experiments.  
More specifically, deep MaE recognition has progressed from analyzing multiple static images to handling complex dynamic image sequences. 
Hereinafter of this section, we provide an in-depth exploration of deep learning-based MaE recognition methods that are tailored for static image analysis (i.e., deep static MaE recognition) and for dynamic sequential image analysis (i.e., deep dynamic MaE recognition). 
By examining the state-of-the-art methodologies, we aim to highlight the strengths and advancements in MaE recognition and to emphasize their potential to address increasingly complex wild scenarios.

\vspace{-0.3 cm}
\subsection{Deep Static MaE Recognition}\vspace{-0.1 cm}
As shown in Table \ref{tb:MaERST}, MaE recognition from multiple static images has been proposed due to the efficiency of data processing and the potential for more detailed analysis.

\vspace{-0.3 cm}
\subsubsection{\bf Ensemble learning} 
Ensemble learning in deep static MaE recognition can exploit complementary information from multiple feature representations of a single emotion image. 
Ensemble learning can be used at different stages of the deep static MaE recognition, such as, 
data preprocessing \cite{Kim_2016_CVPR_Workshops}, 
input enhancement \cite{asd11, HARIRI202184, 9226437, SHAO201982, VISWANATHAREDDY202023}, 
network generation \cite{10.1145/2818346.2830590, 8039231}, 
feature extraction \cite{8371638, 9767587}, and 
emotion classification \cite{9226437,8643924}.
More specifically, deformation and normalization have been used to preprocess the data before training the deep models for MaE recognition \cite{Kim_2016_CVPR_Workshops}. 
When applied at the stage of input enhancement, ensemble learning can improve the MaE recognition by integrating multiple types of textural features, such as, local binary patterns (LBPs) \cite{asd11,HARIRI202184}, facial landmark point \cite{9226437}, covariance feature \cite{SHAO201982}, gradient images \cite{HARIRI202184}, and gravitational force \cite{VISWANATHAREDDY202023}. 
At the network generation stage, deep models with various kernel shapes and parameter initializations are ensembled to improve performance \cite{10.1145/2818346.2830590, 8039231}. 
For example, Saurav et al. \cite{saurav2022dual} proposed an integrated convolutional neural network (CNN) architecture that leveraged two deep models with distinct kernel shapes.
The aggregation of features extracted from deep models represents another prevalent research direction for constructing model ensembles in deep static MaE recognition. 
Several studies \cite{8371638,9767587} have explored integrating different regions (e.g., eyes, mouth, nose, and the entire image) of the face for model ensembling. 
Through the different ensemble strategies \cite{8371638,9767587}, an integrated feature can combine all facial regions for MaE recognition.
Besides the common integration rules (e.g., majority voting, simple average, and weighted average), the weighted average for decision ensemble was also investigated in  \cite{9226437,8643924}. 
Despite the aforementioned merits, ensemble learning increases the training cost and lacks transparency and interoperability.

\vspace{-0.3 cm}
\subsubsection{\bf Transfer learning}  Transfer learning in deep static MaE recognition enables deep models to be fine-tuned on MaE datasets after being pre-trained on large-scale, high-quality related datasets.
The utilization of transfer learning can alleviate the overfitting issue due to the limited training data for static MaE recognition.
By pre-training the model with additional data from other relative tasks, deep static MaE recognition methods can obtain more generic knowledge.
For example, Ng et al. \cite{10.1145/2818346.2830593} used ImageNet \cite{deng2009imagenet} dataset to obtain the pre-trained models that are fine-tuned by the other related MaE datasets.
To further enhance training efficiency, \cite{electronics10091036} fine-tuned only the dense layers of deep models, thereby reducing computational costs.
% The training efficiency can be improved by only fine-tuning the dense layers of deep models; therefore, the training computing expenditure is reduced in \cite{electronics10091036}. 
Instead of using general-purpose datasets, MaE-specific datasets can be used for deep model pre-training \cite{s20092639, Atabansi_2021}. 
For instance, Ngo et al. \cite{s20092639} employed the face identification dataset (e.g., VGGFace2 \cite{8373813}) for model transfer. 
To further mitigate the overfitting in deep static MaE recognition, Atabansi et al. \cite{Atabansi_2021} used another MaE recognition dataset with high resolution and huge quantity for model pre-training.

\vspace{-0.3 cm}
%%%%%%%%%%%%%%%%%%%%%%%%%%%%%%%%%%%%%%%%%%%%%%%%%%%%%%%%%%%%
\subsubsection{\bf Multi-task learning} Multi-task learning allows a deep model to extract macro-expression features from static facial images by using other facial behavior analysis tasks as auxiliary tasks in the deep static MaE recognition.
For example, recent deep static MaE recognition methods have integrated facial landmark localization \cite{CHEN2021107893} and AU detection \cite{pons2018multi} to extract more robust MaE features \cite{pons2018multi, CHEN2021107893, YU2022108401, 10214365}.
Moreover, Pons and Masip found that leveraging shared features across AU and MaE can enhance the performance of recognition \cite{pons2018multi}. 
To further improve the synergy, Li et al. \cite{Li_2023_CVPR} introduced an alignment loss to constrain the feature distribution between the AU detection and MaE recognition tasks.
Furthermore, the multi-task learning can be used to fuse the global and local facial expressions by automatically assigning weights based on the importance of global and local facial information \cite{yu2020facial, XIAO2023110312}.
Besides, facial expression synthesis \cite{9413000}, head pose estimation \cite{chen2022orthogonal}, body gesture \cite{zaghbani2022multi}, and gender learning \cite{9945168} had been demonstrated as promising collaborative auxiliary tasks for deep static MaE recognition and could significantly improve performance.
Therefore, Chen et al. \cite{chen2022orthogonal} proposed a dual-attention based multi-task learning framework that contains a separate channel attention mechanism to calculate task-specific attention weights and an orthogonal channel attention loss to optimize the selection of feature channels for each auxiliary task. 
Since human emotion is conveyed equally via the body and the face in most cases, Zaghbani et al. \cite{zaghbani2022multi} combined the upper body gesture action detection task for the deep static MaE recognition. 
Pan et al. \cite{9945168} incorporated gender learning as an auxiliary task to factor in the effects of gender on the deep static MaE recognition since the characteristics of the same facial expressions from males, females, and infants can be significantly different. 
Instead of using a single auxiliary task, several studies proposed various multiple tasks for deep static MaE recognition to achieve more synergy.
For instance, Qin et al. \cite{10216308} developed an algorithm that simultaneously performs face recognition, MaE recognition, age estimation, and face attribute prediction. 
Similarly, Foggia et al. \cite{FOGGIA2023105651} proposed a comprehensive multi-task framework to integrate gender, age, ethnicity, and MaE recognition using facial images.
	
\vspace{-0.3 cm}
%%%%%%%%%%%%%%%%%%%%%%%%%%%%%%%%%%%%%%%%%%%%%%%%%%%%%%%%%%%%
\subsubsection{\bf Attention-based learning} Attention-based learning is used to enrich the spatial information in deep static MaE recognition. 
The attention modules can focus on both global and local regions and generate more comprehensive static facial expression features that typically include local information, global information, and the corresponding dependency.
Drawing inspiration from the human ability to locate salient objects in complex visual environments, attention mechanisms have emerged as a powerful tool in deep static MaE recognition \cite{itti2001computational}. 
By prioritizing critical facial regions, attention modules can significantly enhance the extraction and integration of expression features in order to achieve a more accurate and robust recognition model.
For example, Fernandez et al. \cite{Fernandez_2019_CVPR_Workshops} developed an attention module to enhance the extraction of expression features by assigning higher weights to relevant regions. 
Li et al. \cite{LI2020340} integrated LBP and deep features with an attention mechanism to enhance the performance of deep static MaE recognition. 
In wild scenarios, pose variations and occlusions present significant challenges for deep static MaE recognition \cite{8576656, wang2020region}. To address the aforementioned issues, various attention modules have been proposed to enhance model robustness and performance.
For example, Li et al. \cite{8576656} introduced an attention module designed to identify local facial regions associated with MaEs while simultaneously incorporating complementary global facial information. 
This dual-focus approach enables more robust MaE recognition under occlusion conditions by balancing localized details with a holistic view of the face.
Wang et al. \cite{wang2020region} proposed a region attention module that is tailored to address occlusions and pose variations. 
The region attention module in \cite{wang2020region} highlights significant facial regions by leveraging attention weights and combines local region features to generate a compact and fixed-length feature representation for final classification.
Zhao et al. \cite{9474949} introduce a global–local attention module with multi‑scale receptive fields.
The global branch gathers broad contextual cues, while the local branch highlights the most salient facial regions.
Together, these two streams yield a richer and more discriminative representation for facial‑expression recognition.
Liu et al. \cite{9750079} design an adaptive, multi‑layer perceptual attention module that weights facial attributes according to human‑perception cues. By jointly emphasizing global context and salient local details, the module markedly improves the robustness of static MaE recognition.

First proposed in \cite{dosovitskiy2020image},  vision transformers (ViTs) have brought remarkable advancements in the domain of deep static MaE recognition, primarily due to their ability to capture long-range dependencies across image patches \cite{aouayeb2021learning, Zheng_2023_ICCV}. By leveraging the self-attention module inherent in transformers, Aouayeb et al. \cite{aouayeb2021learning} pioneered the integration of ViTs with a squeeze-and-excitation block that can extract expression-related features. 
Xue et al. \cite{Xue_2021_ICCV} exploited the ViT architecture to explore relationships among various facial regions for a more holistic understanding of static MaEs.
To address the occlusion issue, Liu et al. \cite{LIU2023781} introduced a patch attention module that assigns attention weights to local facial regions. 
The integration of the patch attention module with a ViT can effectively capture both local and global dependencies. 
Building upon the advancements of ViTs, Zheng et al. \cite{Zheng_2023_ICCV} developed a dual-stream model that leverages a cross-fusion mechanism to combine facial landmark features with image-based features. 
Furthermore, the incorporation of additional modalities has proven instrumental in augmenting transformer-based deep static MaE recognition. 
For example, gradient images \cite{10447031} and gray-level co-occurrence matrices \cite{10061322} have been used as complementary inputs to enrich the feature space. These modalities not only enhance the discriminative power of the models but also provide deeper insights into the complex interplay of facial features.
The optimization of transformer architectures also plays a critical role in advancing deep static MaE recognition.
In \cite{10018959}, Feng et al. demonstrated that adjusting the model structures of ViTs through optimization algorithms can induce significant performance improvements.

\vspace{-0.4 cm}
%%%%%%%%%%%%%%%%%%%%%%%%%%%%%%%%%%%%%%%%%%%%%%%%%%%%%%%%%%%%
\subsubsection{\bf Self-supervised learning} Self-supervised learning can enhance deep static MaE recognition models by broadening their ability to understand and analyze MaEs from a spatial perspective.
More specifically, self-supervised learning is utilized to extract meaningful features from unlabeled data that come from unannotated MaEs \cite{9511468, Wang_2022_ACCV, CHEN2023206, 10121455, an2024self}, data captured from different perspectives \cite{roy2021self, 10.1145/3632960, pmlr-v139-zbontar21a}, and multi-modal data sources \cite{9206016, electronics12020288, halawa2024multi}. 

When applying for unlabeled MaE datasets, Li et al. \cite{9511468} proposed a self-supervised learning method to facilitate compound MaE recognition with multiple labels. 
To avoid the expenditure of manual annotation, Wang et al. \cite{Wang_2022_ACCV} introduced an automatic occluded MaE recognition method that can effectively use large volumes of unlabeled data.
Chen et al. \cite{CHEN2023206} combined self-supervised learning with few-shot strategies to train deep models for static MaE recognition with a limited amount of labeled data.
Fang et al. \cite{10121455} introduced an innovative approach by leveraging contrastive clustering in self-supervised models. 
The proposed method in \cite{10121455} can strategically refine pseudo labels within face recognition datasets and thereby unlock the potential to enhance deep static MaE recognition. 
An et al. \cite{an2024self} proposed a self-supervised static MaE recognition that learns multi-level facial features without requiring labeled data.

The self-supervised learning can be employed for deep static MaE recognition tasks using non-frontal data samples.  
For example, Roy et al. \cite{roy2021self} proposed a contrastive learning method for multi-view MaEs to address the viewpoint sensitivity and limited quantities of labeled data. 
By aligning features of the same expression from different perspectives, the proposed contrastive learning method can generate view-invariant embedding features for multi-view static MaE recognition. 
Later, Roy et al. \cite{10.1145/3632960} introduced an improved version of the proposed contrastive learning method in \cite{roy2021self}. 
After obtaining effective view-invariant features, the proposed approach in \cite{10.1145/3632960} can incorporate supervised contrastive loss and Barlow Twins loss \cite{pmlr-v139-zbontar21a} to further differentiate MaE features with minimized redundancy.

The self-supervised learning has also been applied to multi-modal MaE recognition to address the challenges of integrating data from various modalities without explicit annotation.
Siriwardhana et al. \cite{9206016} utilized pre-trained self-supervised models to extract text, speech, and vision features to improve deep static MaE recognition tasks.
Moreover, multi-modal self-supervised learning frameworks have been explored to extract and fuse multi-modal features to achieve promising results in deep static MaE recognition \cite{electronics12020288, halawa2024multi}.

\vspace{-0.4 cm}
\subsubsection{\bf Integrated Comparisons \& Practical Insights on Deep Static MaE Recognition}
The transfer learning, multi-task learning, and self-supervised learning paradigms can enhance the deep static MaE recognition by feature generalization.  
While enriching feature representation via pre-training (transfer learning \cite{10.1145/2818346.2830593, deng2009imagenet, electronics10091036, s20092639, Atabansi_2021, 8373813}), multi-objective optimization (multi-task learning \cite{CHEN2021107893, pons2018multi, YU2022108401, Li_2023_CVPR, yu2020facial, XIAO2023110312, 9413000, chen2022orthogonal, zaghbani2022multi, 9945168, 10216308, FOGGIA2023105651}), and unlabeled data exploitation (self-supervised learning \cite{9511468, Wang_2022_ACCV, CHEN2023206, 10121455, an2024self, roy2021self, 10.1145/3632960, pmlr-v139-zbontar21a, 9206016, electronics12020288, halawa2024multi}), the aforementioned learning paradigms are still hindered by domain shifts, insufficient labeled data, and intricate training processes. 
Complementary techniques (e.g., ensemble learning \cite{Kim_2016_CVPR_Workshops, asd11, HARIRI202184, 9226437, SHAO201982, VISWANATHAREDDY202023, 10.1145/2818346.2830590, 8039231, 8371638, 9767587, 8643924, saurav2022dual}
and attention-based learning \cite{itti2001computational, Fernandez_2019_CVPR_Workshops, LI2020340, 8576656, wang2020region, 9474949, 9750079, dosovitskiy2020image, aouayeb2021learning, Zheng_2023_ICCV, Xue_2021_ICCV, LIU2023781, 10447031, 10061322, 10018959})
 can further improve accuracy by enhancing feature representation and contextual understanding. 
However, these methods often come with substantial computational costs due to the use of multiple backbones, additional processing pipelines, and more complex model architectures. For a detailed comparison of the strengths and weaknesses, please refer to Appendix 3.1.

A central challenge in deep static MaE recognition lies at balancing accuracy with computational efficiency. 
Complex models usually yield good accuracy. For example, transformer‑based learning can achieve state‑of‑the‑art results by modeling complicated context information \cite{aouayeb2021learning,Zheng_2023_ICCV,Xue_2021_ICCV}.
However, the substantial computing loads are burdensome for the low-power IoT devices.
Two major strategies have been explored, i.e., \emph{model simplification} and \emph{computation offloading}.
For the model simplification, lightweight and pruned networks preserve most of the accuracy while sharply reducing parameters and computation \cite{SHAO201982, 9767587}.
For the computing offloading, inference is (partially) wholly migrated to edge or cloud servers with higher compute capability, and thereby reduce the local latency without sacrificing performance \cite{9767587}.
A second major hurdle is reliable MaE recognition in the wild, where frequent occlusions and pose variations present. 
One effective solution is to construct robust feature representations to occlusions and pose variations. 
This can be achieved through multi-task learning \cite{chen2022orthogonal,zaghbani2022multi,zaghbani2022multi} and self-supervised learning \cite{roy2021self, 10.1145/3632960, pmlr-v139-zbontar21a}, which help the model capture robust, task-related features. 
Additionally, attention-based learning \cite{8576656, wang2020region} can be employed to enhance occlusion awareness by computing attention weights across global and local features, allowing the model to dynamically focus on unoccluded and informative regions of the face.
These strategies push static MaE recognition toward robust, resource‑aware deployment in the wild.

\vspace{-0.35 cm}
	\begin{table}[htb]\scriptsize
	\caption{An Overview of Representative Methods on Deep Dynamic MaE Recognition} \label{tb:MaERDY}
	\vspace{-0.35 cm}
	\resizebox{\textwidth}{!}{
		\renewcommand{\arraystretch}{1.1}
		% \usepackage{multirow}
		\begin{tabular}{|c|l|c|c|c|c|c|c|c|}
			\hline
			Learning Paradigms                                & \multicolumn{1}{c|}{Method} & Year & Pre-processing & Block       & Dataset                                                                         & Performance                                                                                                          & \begin{tabular}[c]{@{}c@{}}Network\\ Structure\end{tabular} & Protocal         \\ \hline
			\multirow{5}{*}{Ensemble Learning}                & Kahou et al. \cite{kahou2013combining}               & 2013 & FA+DA          & N/A         & AFEW2                                                                           & 41.03\%                                                                                                              & CNN                                                         & Cross validation \\ \cline{2-9} 
			& Kahou et al.  \cite{kahou2016emonets}              & 2016 & FA+DA          & N/A         & AFEW4                                                                           & 47.67\%                                                                                                              & CNN                                                         & 2 fold           \\ \cline{2-9} 
			& Liu et al.  \cite{LIU2022182}                & 2022 & DA             & RES+Atten   & BU-3DFE/MMI/AFEW/DFEW                                                           & 85.33\%/91\%/53.98\%/65.35\%                                                                                         & CNN                                                         & Cross validation \\ \cline{2-9} 
			& Liu et al.  \cite{LIU2023109368}                & 2023 & DA             & Trans       & BU-3DFE/MMI/AFEW/DFEW                                                           & 88.17\%/92.5\%/54.26\%/65.85\%                                                                                       & CNN                                                         & Cross validation \\ \cline{2-9} 
			& Pan et al.   \cite{PAN2024120138}               & 2024 & FA             & RES+Atten   & eNTERFACE’05/BAUM-1s/AFEW                                                       & 54.6\%/57.5\%/54.7\%                                                                                                 & CNN                                                         & Cross validation \\ \hline
			\multirow{6}{*}{Explicit Spatio-temporal Learning} & Zhang et al.  \cite{7890464}              & 2017 & DA             & N/A         & CK+/Oulu-CASIA/MMI                                                              & 98.5\%/86.25\%/81.18\%                                                                                                 & CNN+RNN                                                     & 10 fold          \\ \cline{2-9} 
			& Zhi et al.   \cite{8785844}               & 2019 & DA             & N/A         & Biovid Heat Pain/CASME II/SMIC                                                  & 29.7\%/54.6\%/57.8\%                                                                                                 & RNN                                                         & LOGO             \\ \cline{2-9} 
			& Lee et al.  \cite{9102419}                & 2020 & DA             & Atten       & MAVFER                                                                          & RMSE/CC/CCC:0.112,0.563,0.521                                                                                        & RNN+CNN                                                     & Hold out         \\ \cline{2-9} 
			& Hasani et al.  \cite{Hasani_2017_CVPR_Workshops}             & 2017 & FA+DA          & RES         & CK+/MMI/FERA/DISFA                                                              & 93.21\%/77.50\%/77.42\%/58.00\%                                                                                      & CNN+LSTM                                                     & 5 fold           \\ \cline{2-9} 
			& Deng et al.  \cite{8845493}              & 2019 & N/A            & RES         & CK+/MMI/AFEW                                                                    & 94.39\%/80.43\%/82.36\%                                                                                              & CNN+RNN                                                     & 5 fold           \\ \cline{2-9} 
			& Khanna et al.   \cite{khanna2024enhanced}            & 2024 & DA             & RES         & Ravdess/CK+/Baum1                                                               & 91.69\%/98.61\%/73.73\%                                                                                              & CNN                                                         & Hold out         \\ \hline
			\multirow{3}{*}{Multi-task Learning}              & Yu et al.  \cite{yu2020facial}                 & 2020 & FA+FD          & N/A         & CK+/Oulu-CASIA                                                                  & 98.77\%/90.40\%                                                                                                        & CNN+LSTM                                                    & 10 fold          \\ \cline{2-9} 
			& Jin et al. \cite{Jin_2021_ICCV}                 & 2021 & DA             & RES         & Aff-wild2                                                                       & 77.09\%                                                                                                              & CNN                                                         & Hold out         \\ \cline{2-9} 
			& Xie et al.   \cite{XIE2023126649}               & 2023 & FA+DA+ROI      & Trans+Atten & CMU-MOSI/CMU-MOSEI                                                               & 85.36\%/84.61\%                                                                                                      & BERT                                                        & Hold out         \\ \hline
			\multirow{13}{*}{Attention-Based Learning}        & Meng et al.  \cite{8803603}               & 2019 & FD+FA          & RES+Atten   & CK+/AFEW                                                                    & 99.69\%/51.18\%                                                                                                       & CNN                                                         & 10 fold          \\ \cline{2-9} 
			& Liu et al.  \cite{LIU2020145}                & 2020 & FD+DA          & Atten       & CK+/Oulu-CASIA/MMI/AffectNet                                                    & 99.54\%/88.33\%/87.06\%/63.71\%                                                                                      & CNN+RNN                                                     & 10 fold          \\ \cline{2-9} 
			& Sun et al.  \cite{SUN2021378}                & 2021 & FD             & Atten       & CK+/MMI /Oulu-CASIA                                                             & 99.1\%/89.88\%/87.33\%                                                                                               & 2sCNN                                                       & 10 fold          \\ \cline{2-9} 
			& Xia et al. \cite{xia2022multi}                 & 2022 & FD             & Atten       & Aff-Wild2/RML/AFEW                                                              & 50.3\%/78.32\%/59.79\%                                                                                               & CNN+3DCNN                                                   & Hold out + LOSO  \\ \cline{2-9} 
			& Chen et al.  \cite{9209166}               & 2020 & FD+DA          & Atten       & CK+/Oulu-CASIA/MMI                                                              & 99.08\%/91.25\%/82.21\%                                                                                              & 3D CNN                                                      & 10 fold          \\ \cline{2-9} 
			& Zhao et al.  \cite{10.1145/3474085.3475292}               & 2021 & FD+DA          & RES+Atten   & DFEW/AFEW                                                                       & \begin{tabular}[c]{@{}c@{}}UAR: 53.69\%/47.42\%\\ WAR:65.70\%/50.92\%\end{tabular}                                       & CNN                                                         & 5 fold+Hold out  \\ \cline{2-9} 
			& Huang, et al.  \cite{HUANG202135}             & 2021 & FA             & RES+Trans   & CK+/FERplus/RAF-DB                                                              & 100\%/90.04\%/88.26\%                                                                                                & CNN                                                         & 10 fold+Hold out \\ \cline{2-9} 
			& Ma et al.  \cite{ma2022spatio}                 & 2022 & FD+FA          & RES+Trans   & DFEW/AFEW                                                                       & \begin{tabular}[c]{@{}c@{}}UAR: 54.58\%/49.11\%\\ WAR: 66.65\%/54.23\%\end{tabular}                                  & CNN                                                         & 5 fold+Hold out  \\ \cline{2-9} 
			& Zhang et al. \cite{Zhang_2022_CVPR}               & 2022 & FD             & RES+Trans   & ABAW                                                                            & F1: 35.9\%                                                                                                           & CNN                                                         & Hold out         \\ \cline{2-9} 
			& Zhao et al. \cite{9794419}                & 2022 & FA+FD          & Atten+Trans & \begin{tabular}[c]{@{}c@{}}CK+/Oulu-CASIA/eNTERFACE05/\\ AFEW/CAER\end{tabular} & \begin{tabular}[c]{@{}c@{}}98.78\%/89.17\%/54.62\%/\\ 51.17\%/77.06\%\end{tabular}                                   & GCN                                                         & Cross validation \\ \cline{2-9} 
			& Wang et al. \cite{10446699}                & 2024 & FA             & RES+Trans   & DFEW/FERV39k                                                                    & \begin{tabular}[c]{@{}c@{}}UAR: 60.28\%/41.28\%\\ WAR: 71.42\%/51.02\%\end{tabular}                                  & CNN                                                         & N/A              \\ \cline{2-9} 
			& Chen et al. \cite{CHEN2024123635}                & 2024 & DA             & RES+Trans   & DFEW/FERV39k/AFEW                                                           & \begin{tabular}[c]{@{}c@{}}UAR: 58.65\%/41.91\%/52.23\%\\ WAR: 69.91\%/50.76\%/55.40\%\end{tabular}                  & CNN                                                         & Hold out+5 fold  \\ \cline{2-9} 
			& Zhang et al.  \cite{10250883}              & 2023 & FD+DA          & Trans       & MAFW11/MAFW43/DFEW/AFEW                                                        & \begin{tabular}[c]{@{}c@{}}UAR: 39.37\%/15.22\%/57.16\%/50.22\%\\ WAR: 52.85\%/39.00\%/68.85\% /52.96\%\end{tabular} & CNN+Trans                                                   & 5 fold+Hold out  \\ \hline
			\multirow{3}{*}{Self-supervised Learning}         & Song et al. \cite{9412942}                & 2021 & FD+ROI         & N/A         & SEMAINE                                                                         & MSE:0.058/0.072                                                                                                      & UNet                                                        & Cross validation \\ \cline{2-9} 
			& Sun et al. \cite{10.1145/3581783.3612365}                 & 2023 & ROI            & Trans       & DFEW/FERV39k/MAFW                                                               & \begin{tabular}[c]{@{}c@{}}UAR:63.41\%/43.12\%/41.62\%\\ WAR:74.43\%/52.07\%/54.31\%\end{tabular}                    & Trans                                                       & Cross validation \\ \cline{2-9} 
			& Chumachenko et al. \cite{chumachenko2024mma}         & 2024 & FD             & Trans       & DFEW/MAFW                                                                       & \begin{tabular}[c]{@{}c@{}}UAR:66.85\%/44.25\%\\ WAR:77.43\%/58.45\%\end{tabular}                                    & Trans                                                       & 5 fold           \\ \hline
	\end{tabular}}
	\begin{tablenotes}\scriptsize
		\item $ ^1$DA: Data augmentation; FD: Face detection; CE: Contrast enhancement; FA: Face alignment; N/A: Not applicable; UAR: Unweighted average recall; WAR: Weighted average recall; RMSE: Root mean squared error; CC: Correlation coefficient; CCC: Concordance correlation coefficient.
		\item $ ^2$Trans: Transformer; Atten: Attention; 
		nsCNN: n-stream CNN; LOSO: Leave-one-subject-out; RNN: Recurrent neural network; LSTM: Long short-term memory; BERT: Bidirectional encoder representations from transformer; GCN: Graph convolutional network.
	\end{tablenotes}
\vspace{-0.6 cm}
\end{table}

\vspace{-0.2 cm}
\subsection{Deep Dynamic MaE Recognition}\vspace{-0.1 cm}
While MaE recognition can be achieved using static images and isolated frames, it greatly benefits from analyzing consecutive dynamic frames over time. In this section, we provide a comprehensive review of recent deep dynamic MaE recognition methods, summarized in Table \ref{tb:MaERDY}.

\vspace{-0.3 cm}
\subsubsection{\bf Ensemble learning} Ensemble learning can enhance the accuracy and robustness of dynamic MaE recognition by frame aggregation that can combine temporal dynamic features across multiple frames and frame-level emotion classification results.
In deep dynamic MaE recognition, the frame aggregation can be implemented at decision-level \cite{kahou2013combining, kahou2016emonets} and feature-level \cite{8863974, LIU2022182, LIU2023109368, PAN2024120138}. 

The decision-level frame aggregation integrates the classification results of individual frames in the form of class probability vectors. 
For instance,  Kahou et al. \cite{kahou2013combining} explored the decision-level frame aggregation by using averaging and expansion for deep dynamic MaE recognition. 
Later, Kahou et al. \cite{kahou2016emonets} used expansion or contraction methods to aggregate single-frame probabilities into fixed-length video descriptors.
Since the number of frames may vary, statistical characteristics (e.g., mean, variance, minimum, and maximum) can be utilized to summarize the frame-level outputs. 
However, relying solely on per-frame decisions may overlook the temporal dependencies between consecutive frames, which may impact the performance of deep dynamic MaE recognition.

The feature-level frame aggregation focuses on integrating frame-level features for final prediction. 
Nguyen et al. \cite{8863974} concatenated frame features and fed them into a 3D CNN model for MaE recognition. 
By leveraging the attention mechanism, Liu et al. \cite{LIU2022182} introduced a frame aggregation method to integrate expression-related features. 
Specifically, they segmented videos into short clips and employed an attention-based feature extractor to capture salient features from these segments.
Then, an emotional intensity activation network was designed to locate salient clips for generating robust features.
Building on this, Liu et al. \cite{LIU2023109368} proposed a transformer-based frame aggregation method for dynamic MaE recognition.
To effectively integrate interrelationships among multi-cue features, Pan et al. \cite{PAN2024120138} developed a hybrid fusion method for video-based MaE recognition. 
The proposed ensemble method in \cite{PAN2024120138} combines the strengths of different types of features (e.g., appearance, geometry, and high-level semantic knowledge) and thereby improves the overall performance and robustness of dynamic MaE recognition systems.

\vspace{-0.3 cm}
\subsubsection{\bf Explicit spatio-temporal learning} Explicit spatio-temporal learning focuses on constructing deep models that are designed to extract spatio-temporal information from the dynamic MaE datasets. 
The explicit spatio-temporal learning methods can incorporate the temporal information into the encoded features by using a sequence of frames within a sliding window. 
The recurrent neural network (RNN) \cite{medsker2001recurrent} and its advanced version (i.e., long short-term memory, LSTM)  \cite{hochreiter1997long} have been used in dynamic MaE recognition due to the ability to process sequential data 
\cite{7890464, 9102419, 9134869, 9330541, 10.3745/JIPS.01.0067, 9206081, YU201850, an2020facial, 9051332}. 
For instance, Zhang et al. \cite{7890464} employed an RNN to articulate morphological changes and dynamic evolution of MaEs by exploiting key facial regions based on facial landmarks. 
Zhi et al. \cite{8785844} proposed a lightweight LSTM-based method with less complexity to produce a sparse representation for dynamic MaE recognition. 
Lee et al. \cite{9102419} introduced attention-guided convolutional LSTM to capture dynamic spatio-temporal information.
By leveraging the depth and thermal sequences as guidance priors, the proposed model can guide the model to focus on discriminative visual regions.
An LSTM autoencoder was used to learn temporal dynamics for dynamic MaE recognition \cite{9330541}.
Besides, the 3D deep models (e.g., 3D CNN \cite{ji20123d}) have also been adopted to extract spatio-temporal features directly for dynamic MaE recognition \cite{Hasani_2017_CVPR_Workshops, 8845493, 8373844}. 
Hasani et al. \cite{Hasani_2017_CVPR_Workshops} combined 3D Inception-ResNets \cite{szegedy2017inception} with an LSTM unit to capture spatio-temporal information. 
Deng et al. \cite{8845493} proposed a 3D CNN framework consisting of a stem layer, 3D Inception-ResNets structure, and gated recurrent unit (GRU) layer to process video data. 
To further improve the performance of the model, the island loss \cite{8373844} was incorporated to increase inter-class differences while minimizing intra-class variations.
Khanna et al. \cite{khanna2024enhanced} developed an end-to-end spatio-temporal deep model that integrates residual networks \cite{He_2016_CVPR} and DenseNet \cite{Huang_2017_CVPR} for MaE recognition.

\vspace{-0.3 cm}
\subsubsection{\bf Multi-task learning} Multi-task learning in deep dynamic MaE recognition focuses on capturing the dynamic variation patterns of MaEs over time in order to enhance the understanding of expression changes in video clips.
By fusing local dynamic features extracted from a part-based module, Yu et al. \cite{yu2020facial} proposed a multi-task module that can capture the subtle variations of MaEs from both local and global features.
Jin et al. \cite{Jin_2021_ICCV} utilized a multi-task framework involving AU detection and MaE recognition to pre-train the visual feature extractor. 
Then, both visual and audio features from videos were concatenated and fed into a transformer encoder to extract temporal features for recognition.
Xie et al. \cite{XIE2023126649} expanded the scope of auxiliary tasks by incorporating dynamic MaE recognition from multiple modalities, such as text, audio, and video. 
	
\vspace{-0.3 cm}
\subsubsection{\bf Attention-based learning} Attention-based learning in deep dynamic MaE recognition focuses on both spatial and temporal information. 
Compared with its deep static variants, attention modules in deep dynamic MaE recognition operate across the temporal and even channel dimensions to extract richer video features. 
Furthermore, the attention modules can integrate multimodal features to derive more comprehensive spatio-temporal representations.

Various attention modules have been proposed to enhance deep dynamic MaE recognition by integrating spatial, temporal, and channel-wise information, e.g., self-attention \cite{8803603}, relation attention \cite{8803603}, AU attention \cite{LIU2020145}, and multi-attention \cite{SUN2021378, xia2022multi, 9209166}.
For example, Meng et al. \cite{8803603} proposed a self-attention module to aggregate all input frame features and a relation attention module to capture informative features from global and local contexts.
Liu et al. \cite{LIU2020145} introduced an AU attention mechanism to augment long-range expression information by focusing on specific AU regions.
Besides, multi-attention is implemented in \cite{SUN2021378, xia2022multi, 9209166} to obtain the rich complementary information in deep dynamic MaE recognition.
For instance, Sun et al. \cite{SUN2021378} proposed two attention modules to integrate facial features extracted from two deep models. Subsequently, an additional attention-based deep model was introduced to compose a comprehensive feature representation from MaE sequences.
Xia et al. \cite{xia2022multi} developed a multi-attention module for dynamic MaE recognition. The proposed module in \cite{xia2022multi} can integrate spatial and temporal information from areas of interest (e.g., expressive frames and salient facial patches).
Chen et al. \cite{9209166} further expanded the scope by incorporating channel-wise attention and demonstrated its effectiveness in enhancing both performance and interpretability. 
Specifically, the proposed method in  \cite{9209166} employed a spatio-temporal module and a channel attention module to explore correlations across spatio-temporal and channel dimensions. 
Moreover, the proposed method in  \cite{9209166} can generate spatio-temporal attention maps for visualization.

Transformers have been used to enhance dynamic MaE recognition by integrating spatial, temporal, and contextual video features \cite{10.1145/3474085.3475292, HUANG202135, ma2022spatio, Zhang_2022_CVPR}.
Zhao et al. \cite{10.1145/3474085.3475292} proposed a dual-transformer framework that extracts robust MaE features by leveraging a spatial transformer module to capture spatial configurations and a temporal transformer module to account for temporal changes in expression videos.
Huang et al. \cite{HUANG202135} developed a MaE recognition framework that utilizes grid-wise attention to capture dependencies among facial regions and a ViT module to extract global features.
To address long-range dependencies within videos, Ma et al. \cite{ma2022spatio} designed a spatio-temporal transformer that integrates contextual relationships across frames, providing a comprehensive representation of dynamic MaE features.
Additionally, some studies \cite{9794419,10446699, CHEN2024123635} incorporated external knowledge into transformers to enhance feature extraction. For instance, Zhao et al. \cite{9794419} employed facial landmarks to construct spatio-temporal graphs, using graph convolutional blocks and transformer modules to extract MaE-related features. Wang et al. \cite{10446699} combined multi-scale spatial information from CNNs with temporal features extracted from transformer modules. Similarly, Chen et al. \cite{CHEN2024123635} introduced multi-geometry knowledge into transformers, enabling the extraction of spatio-temporal geometry information for MaE recognition.
	
Moreover, transformers have been leveraged to integrate multi-modal features for dynamic MaE recognition. 
For instance, Zhang et al. \cite{Zhang_2022_CVPR} proposed a transformer-based framework that fuses static visual features with dynamic multi-modal information derived from text, audio, and video. 
In the proposed framework of \cite{Zhang_2022_CVPR}, static frame features act as a guiding mechanism for the multi-modal fusion process to ensure the alignment of temporal dynamics with spatial cues from individual frames.
Building on the framework in \cite{Zhang_2022_CVPR}, Zhang et al. \cite{10250883} introduced an advanced multi-modal transformer framework with improved methods for feature fusion and extraction. Specifically, the framework employs three transformer-based encoders, each dedicated to extracting features from a specific modality—visual, audio, or text—while enabling modality-specific adaptation.
Unlike the earlier framework \cite{Zhang_2022_CVPR} that focused on integrating multi-modal data using static frame guidance, the subsequent method \cite{10250883} achieves more robust multi-modal feature integration by aligning semantic information across modalities by mapping features from all modalities into a shared latent space anchored in visual features, thereby supporting a more unified representation of multi-modal data.

\vspace{-0.5 cm}
\subsubsection{\bf Self-supervised learning} Self-supervised learning focuses on capturing changes in expressions across spatial and temporal dimensions in the deep dynamic MaE recognition. Moreover, self-supervised models can learn spatio-temporal features from unlabeled video data. 
Since the self-supervised learning models can interpret videos by learning the order of frames in human action recognition \cite{Fernando_2017_CVPR, 8756609},  Song et al. \cite{9412942} introduced a self-supervised framework with a rank loss mechanism to generate dynamic MaE features. 
By sorting preceding and following frames based on their distance from a central frame, the proposed framework can generate robust dynamic MaE features.
Sun et al. \cite{10.1145/3581783.3612365} proposed a large-scale self-supervised model for pre-training a dynamic MaE recognition system using extensive unlabeled facial video data. 
Chumachenko et al. \cite{chumachenko2024mma} proposed a multi-modal approach that combined complementary features from diverse modalities. 
The multi-modal approach employed two self-supervised learning encoders pre-trained on audio sequences and static images, and then fine-tuned using audio-visual dynamic videos.

\vspace{-0.3 cm}
\subsubsection{\bf Integrated Comparisons \& Practical Insights on Deep Dynamic MaE Recognition}
Deep dynamic MaE recognition enables models to capture the temporal evolution of emotions by expanding facial expression analysis to video sequences.
Various learning paradigms have been proposed to extract spatiotemporal features critical for accurate dynamic MaE recognition.
Recent studies explore a range of strategies, including frame-level integration  (ensemble learning \cite{kahou2013combining, kahou2016emonets, 8863974, LIU2022182, LIU2023109368, PAN2024120138}), modeling spatio-temporal dynamics (explicit spatio-temporal learning \cite{medsker2001recurrent, hochreiter1997long, 7890464, 9102419, 9134869, 9330541, 10.3745/JIPS.01.0067, 9206081, YU201850, an2020facial, 9051332, 8785844, ji20123d, Hasani_2017_CVPR_Workshops, 8845493, 8373844, szegedy2017inception, khanna2024enhanced, He_2016_CVPR, Huang_2017_CVPR}, attention‑based learning \cite{8803603, LIU2020145, SUN2021378, xia2022multi, 9209166, 10.1145/3474085.3475292, HUANG202135, ma2022spatio, Zhang_2022_CVPR, 9794419, 10446699, CHEN2024123635, 10250883}), and fusing multi-source features (multi-task learning \cite{yu2020facial, Jin_2021_ICCV, XIE2023126649}) for improved dynamic MaE recognition.
However, these approaches still face challenges such as the lack of explicit temporal dependency modeling, high training complexity, and the scarcity of well-aligned benchmark datasets.
Self-supervised learning \cite{Fernando_2017_CVPR, 8756609, 9412942, 10.1145/3581783.3612365, chumachenko2024mma} partially mitigates the data‑scarcity issue by exploiting unlabeled videos, but the cost of lengthy training cycles and substantial compute requirements limits its practical deployment.
A detailed comparison of deep dynamic MaE recognition methods is provided in Appendix 3.2.

The temporal nature of dynamic MaE recognition introduces unique bottlenecks.
For example, traditional 3D CNNs that process entire video clips for analysis can increased latency, which is a critical issue for edge devices handling real-time streams.
One effective solution is to reduce the complexity of the input representation. 
For example, \cite{8845493} leverage facial landmarks to extract temporal MaE cues while integrating spatial appearance features from individual frames. 
Similarly, \cite{9794419} demonstrate that expressive spatiotemporal dynamics can be captured using only landmark sequences, bypassing raw pixel input.
In addition, \cite{9412942} further show that dynamic representations can even be derived from static images, supporting the feasibility of key-frame-based recognition pipelines.
An alternative solution focuses on architectural efficiency. Recent approaches \cite{10446699, CHEN2024123635} adopt lightweight backbones such as ResNet-18 for spatial encoding, coupled with transformer-based temporal modules to capture global temporal dependencies. 
This modular design enables flexible control over the number of input frames, providing a practical trade-off between latency and recognition performance.

Another major challenge in dynamic MaE recognition is the scarcity of large-scale labeled datasets. 
Existing datasets used for practical applications(e.g., DFEW, AFEW) are limited in size, which hinders the the training of deep temporal models with strong generalization capabilities.
A widely adopted solution to this problem is the use of self-supervised pre-training on large-scale unlabeled but domain-relevant data. 
This approach enables models to learn generalized MaE representations, which can be fine-tuned for downstream dynamic facial expression recognition tasks.
For instance, \cite{10.1145/3581783.3612365, chumachenko2024mma} demonstrate that applying a masked autoencoding reconstruction objective on unlabeled MaE data can enhance performance on dynamic recognition tasks.
In addition, \cite{9412942} show that learning temporal ordering between adjacent frames provides valuable temporal supervision, allowing the model to capture dynamic dependencies and improve its ability to model subtle emotional transitions across time.

\vspace{-0.2 cm}
\section{Holographic MiE Analysis}\label{commier}\vspace{-0.2 cm}
Detecting MiEs is challenging due to their short duration and subtle emotional cues; therefore, each MiE must first be identified before proceeding to recognition. 
As a foundational step, MiE spotting involves identifying the temporal and spatial occurrences of these brief and subtle MiEs within a video clip. 
Building on MiE spotting, MiE recognition then analyzes and classifies the detected MiEs.
In the remainder of this section, we detail the state-of-the-art approaches to the two major steps of holographic MiE analysis, i.e. MiE spotting and MiE recognition.

\vspace{-0.4 cm}
\subsection{MiE Spotting}\vspace{-0.1 cm}
MiE spotting, referring to locating temporal segments in a specific video clip, is an important step for holographic MiE analysis. 
The key description of MiE in a video clip is to identify three time spots: (1) the onset refers to the initial frame where the emotion of a MiE becomes discernible; (2) the apex frame is the frame within the MiE clip that exhibits the highest intensity of emotion; and (3) the offset frame marks the end of the emotion. 
The current works on MiE spotting can be categorized into apex-frame spotting and interval spotting. 
%	The process of MiE spotting is the initial step in various analysis tasks, as it can identify the specific time spots in a video where MiEs occur. 

\vspace{-0.1 cm}	
\subsubsection{\bf Apex frame spotting}\vspace{-0.1 cm}
The apex frame reflects the highest intensity of a MiE within a video clip and implies the ground-truth emotion of individuals. 
Traditional apex frame spotting methods exploited the algorithms based on handcrafted features, such as the descriptors of LBP and its variants due to their effectiveness and robustness. 
Pfister et al. \cite{pfister2011recognising} pioneered a MiE spotting approach for short spontaneous expressions.
The proposed method in \cite{pfister2011recognising} leverages local binary patterns from three orthogonal planes (LBP-TOP) \cite{zhao2007dynamic} to achieve notable results compared to manual detection methods. 
Yan et al. \cite{yan2015quantifying} proposed to utilize two handcrafted descriptors (i.e., the constrained local model \cite{cristinacce2006feature} and LBP) to detect the ROI of each face and to extract subsequent features for MiE analysis.
To address the limitations of LBP-TOP in capturing essential information, Esmaeili et al. \cite{esmaeili2020automatic} proposed Cubic-LBP---an improved version of LBP.
Compared with LBP, Cubic-LBP can extract fine-grained features from 15 planes and integrate the obtained histograms. 
Although the use of additional planes could enrich the features, Cubic-LBP also introduced redundant data that increases computational expenditure.
To alleviate the computational expenditure, Esmaeili et al. \cite{esmaeili2022spotting} proposed an intelligent Cubic-LBP by leveraging CNNs to select relevant planes.

MiE spotting can be advanced through frequency-domain features \cite{li2018can, li2020joint}, CNN-based sliding windows \cite{zhang2018smeconvnet}, and attention mechanisms \cite{yee2022apex} for enhanced performance and feature selection.
Given the challenges in capturing MiE motion within the spatio-temporal domain, an alternative approach is to extract features from the frequency domain. 
Li et al. \cite{li2018can, li2020joint} introduced a spotting method leveraging the frequency domain to identify the apex frame.
The proposed methods in \cite{li2018can, li2020joint} demonstrated the effectiveness of the frequency domain feature in MiE spotting. 
The CNN models have also gained traction in MiE spotting. 
Zhang et al. \cite{zhang2018smeconvnet} first employed CNNs with a sliding window approach to locate apex frames in long videos. 
To spot apex frames in onset-offset temporal sequences, Yee et al. \cite{yee2022apex} used an attention module to highlight key regions in optical flow. 
The proposed method in \cite{yee2022apex} underscores that the attention mechanism has the capability of automatic feature selection in respective regions; therefore, the spotting performance can be improved.

\vspace{-0.1 cm}	
\subsubsection{\bf Interval spotting} \vspace{-0.1 cm}	
Interval spotting refers to detecting the onset and offset in a long MiE video with various constraints, such as, MaE, head movement, and blinks. 
While more challenging than apex frame spotting, interval spotting has broader practical applications.
For interval spotting, traditional methods typically set thresholds to locate key points by comparing feature differences across entire videos. 
For example, Li et al. \cite{li2017towards} proposed a method based on comparison of difference characteristics.
Specifically, the proposed method identified LBP as a more effective baseline technique compared to the optical flow histogram.
By incorporating the multi-scale analysis and a sliding window, Tran et al. \cite{tran2017sliding} scaled the video sequence and obtained corresponding samples for spotting MiE intervals. 
Optical flow has been used in interval spotting due to its ability to describe motion information.
Patel et al. \cite{patel2015spatiotemporal} introduced a heuristic algorithm with optical flow for MiE interval spotting.
The algorithm emphasized the continuity of motion direction. 
By analyzing the direction of motion vectors over time, it ensured that detected movements extracted by optical flow were consistent. 
Wang et al. \cite{wang2016main,wang2017main} introduced the main directional maximal difference (MDMD) to characterize facial motion by calculating the largest magnitude difference in the main direction of optical flow. 
The proposed method was later adopted as a baseline in the third facial MiE grand challenge \cite{jingting2020megc2020}. 
Han et al. \cite{han2018cfd} enhanced feature difference analysis by combining LBP with the main directional mean optical flow (MDMO) \cite{liu2015main}, which leveraged both texture and motion features for complementary insights. 
Guo et al. \cite{guo2021magnitude} extended this work by incorporating both magnitude and angle information from optical flow to detect local MiE movements.

MiE interval spotting methods focusing on handcrafted features and optical flow mentioned above require appropriate threshold settings, which is susceptible to the selected features and unexpected facial motion. 
To address the challenges, deep model approaches have emerged as promising alternatives.
To distinguish the MaE and MiE in a long video sequence, Pan et al. \cite{pan2020local} presented a local bilinear CNN to detect the fine-grained facial regions associated with MiE.
Takalkar et al. \cite{takalkar2021lgattnet} developed LGAttNet, a dual-attention framework that integrated local and global facial features for improved spotting. 
Gu et al. \cite{gu2023lite} proposed a lightweight deep model to predict the likelihood of a frame being part of a MiE interval for overfitting mitigation.
To capture spatio-temporal features, Wang et al. \cite{wang2021mesnet} designed a 2D spatial and 1D temporal convolutional model. 
In addition, many studies \cite{yu2021lssnet,yap20223d} utilized 3D CNNs to extract robust spatio-temporal features. 
Zhou et al. \cite{zhou2022novel} further enhanced performance by integrating 3D CNNs with bidirectional encoder representations from transformers (BERTs) for onset-offset detection. 
Moreover, approaches leveraging AU analysis \cite{yin2023aware,yang2023deep} can incorporate prior knowledge and reduce noise interference to improve MiE spotting.
	
\vspace{-0.3 cm}
\subsubsection{\bf Integrated Comparisons \& Practical Insights on MiE spotting}
MiE spotting is a precursor to successful MiE recognition, which demands precise temporal sensitivity to localize transient facial cues often embedded within long, expression-neutral sequences.
Early methods relied heavily on handcrafted motion descriptors, such as optical flow \cite{patel2015spatiotemporal,wang2016main,wang2017main} and LBP on temporal planes \cite{pfister2011recognising,esmaeili2020automatic}, while more recent approaches have adopted deep learning techniques that encode temporal patterns from full video sequences or sparse frames \cite{zhang2018smeconvnet,yee2022apex,pan2020local,takalkar2021lgattnet,gu2023lite,wang2021mesnet,yu2021lssnet,yap20223d,zhou2022novel,yin2023aware,yang2023deep}. 
These methods attempt to balance the trade-off between computational cost, robustness to noise, and the fidelity of motion representation. Despite advancements, spotting remains an underexplored and error-prone stage due to ambiguous expression boundaries and individual-specific motion variances. 
A detailed comparison of deep dynamic MaE recognition methods is provided in Appendix 4.1.

Despite a decade of progress, reliable MiE spotting remains elusive in real‑world settings. 
Three interrelated issues are highlighted below.
Lack of a unified evaluation protocol. Current studies report results under heterogeneous conditions, which hinders fair comparison. For apex spotting, some authors process isolated MiE clips whereas others search for apex frames in long, mixed‑expression videos, leading to incomparable recall and precision numbers. Interval‑level spotting has likewise evolved from early ``enlarged-window'' hit tests \cite{li2017towards} to the now popular Intersection‑over‑Union (IoU $\geq$ 0.5) criterion \cite{pan2020local}. A consensus benchmark is urgently needed: (1) apex spotting should be evaluated in long, unconstrained sequences that include distractor expressions; and (2) interval spotting should adopt IoU‑based metrics, which align with modern object‑detection practice.
Sensitivity to illumination and head motion. Hand‑crafted descriptors encode pixel‑wise differences and optical flow, making them brittle under pose changes and lighting fluctuations. Li et al.~\cite{li2017towards}, for instance, report frequent false positives caused by eye blinks. Moreover, their pipeline requires a temporal interpolation model that alone costs $\approx$22 s per video, pushing total inference time beyond usability for online applications.
Limited recall in deep models for long videos. Deep architectures overcome the temporal interpolation model bottleneck by feeding fixed‑stride windows or variable‑length clips directly to 3‑D CNNs, temporal transformers, or graph–temporal hybrids. Although this reduces preprocessing latency and boosts interval‑level F1 scores (e.g., 38.34\% on CAS(ME)\textsuperscript{2} and 37.28\% on SAMM‑LV \cite{yin2023aware}), performance drops when focusing exclusively on MiE (15.38\% and 19.84\%, respectively). These results indicate that current spotting models are still insufficiently sensitive to the sparse, low‑intensity cues that distinguish MiEs from surrounding MaEs.

\vspace{-0.3 cm}
\subsection{\bf MiE Recognition}\label{mierec}\vspace{-0.1 cm}
Since the seminar work \cite{pfister2011differentiating}, MiE recognition has seen unprecedented development that ranges from shallow machine learning with handcrafted descriptors to deep learning. 
Hereinafter, the MiE recognition will be thoroughly reviewed. 
	
\vspace{-0.3 cm}
\subsubsection{\bf Shallow machine learning methods}
The success of shallow machine learning\footnote{The word ``shallow'' is used to differentiate from the deep learning methods.}  demands for well-designed handcrafted descriptors (e.g., LBPs and grayscale images) to extract local and global facial information. 
As a key component of MiE recognition, the development of descriptors has attracted several research endeavors, such as, LBP with six intersection points \cite{wang2015lbp} and LBP with three mean orthogonal planes \cite{wang2015efficient}.
More specifically, 	Wang et al. \cite{wang2015lbp} proposed the LBP with six intersection points to reduce the redundancy inherent in LBP-TOP. 
Later, Wang et al. \cite{wang2015efficient} proposed the LBP with three mean orthogonal planes to alleviate the computational complexity issue of LBP-TOP.
Compared with the LBP-TOP descriptor, the proposed LBP with three mean orthogonal planes offers faster processing with a competitive performance \cite{wang2015efficient}.
Huang et al. \cite{huang2016spontaneous} developed the spatio-temporal completed local quantization pattern descriptor that incorporates additional information on sign, magnitude, and orientation components to enhance MiE recognition. 
To address the limitation of LBP-TOP on capturing muscle movement in the oblique direction, Wei et al. \cite{wei2022micro} proposed a descriptor named as LBP with five intersecting planes (LBP-FIP).
The LBP-FIP descriptor enhances the LBP-TOP with an innovative feature (i.e., eight vertices LBP) in order to enrich the feature information by capturing detailed MiE dynamics in multiple directions.
Therefore, the LBP-FIP can provide a richer and more effective representation of MiE recognition than the LBP-TOP descriptor.

Besides the series of LBP descriptors, the histogram of gradient (HOG) emerged as another prominent descriptor for MiE recognition \cite{dalal2005histograms}.
By computing and aggregating the gradient directions in local regions of a facial image, HOG demonstrates its robustness to variations in illumination and deformation. 
Therefore, various HOG-based descriptors were adapted for MiE recognition \cite{li2017towards, niu2018micro, zhang2020new}.
Li et al. \cite{li2017towards} proposed the histogram of image gradient orientation, which further suppressed illumination issues by discarding magnitude weighting from the first-order derivative. 
Niu et al. \cite{niu2018micro} introduced a local second-order gradient pattern to capture subtle facial changes in brief video clips. 
Zhang et al. \cite{zhang2020new} leveraged the histograms of oriented gradients on three orthogonal planes and the histograms of image-oriented gradients on three orthogonal planes to construct an effective MiE recognition framework.

Due to the superiority of describing brightness patterns in images, optical flow \cite{horn1981determining} can also be used to capture subtle muscle motion from adjective frames for MiE recognition.
More specifically, Liu et al. \cite{liu2015main} introduced the MDMO feature descriptor for MiE recognition.
MDMO feature descriptor can integrate local motion information and spatial location from partitioned ROIs on the face, providing a robust representation of facial movements.
Liong et al. \cite{liong2014optical} proposed to use the optical strain,  consisting of the shear and normal strain tensors from optical flow, for MiE recognition.
Happy et al. \cite{happy2017fuzzy, happy2018recognizing} developed a fuzzy histogram of optical flow orientation to encode the temporal patterns of facial micro-movements; therefore, distinctive features that were sensitive to the nuances of MiE can be extracted. 
Considering the directional continuity of facial motion, Patel et al. \cite{patel2015spatiotemporal} proposed an optical flow-based method that extracts features from local spatial regions with spatio-temporal integration.

Besides the feature extraction strategies, several works have investigated feature selection \cite{happy2018recognizing, wang2022fixed} and feature fusion \cite{yu2018spatiotemporal, zhang2020new} for MiE recognition. 
By selecting a subset of relevant features, feature selection enhances the MiE recognition methods by reducing the redundant high-dimensional features.
Happy et al. \cite{happy2018recognizing} utilized a fuzzy histogram of optical flow orientation descriptors to capture temporal patterns associated with facial micro-movements. 
Then, the proposed method explores various feature selection methods to reduce the dimension of feature space.
To extract effective spatio-temporal features, Yu et al. \cite{yu2018spatiotemporal} combined the improved local directional number pattern in the spatial domain with the pyramid of histograms of orientation gradients in the temporal domain.
Zhang et al. \cite{zhang2020new} proposed a feature selection method to integrate the effective feature components from the seven local regions of the face.
Wang et al. \cite{wang2022fixed} introduced an integral projection method to enhance information density, followed by a fixed-point rotation-based feature selection approach to identify features with significant motion variations. 
To address redundant or misleading features in recognition tasks, Wei et al. \cite{wei2022learning} implemented a kernelized two-group sparse learning model to optimize feature discrimination.

In general, shallow machine learning methods are based on handcrafted descriptors that demand extensive prior knowledge on MiEs.  
After extracting discriminative features, classifying methods (e.g., support vector machine \cite{yu2018spatiotemporal} and $k$-nearest neighbor \cite{happy2018recognizing}) are used for final MiE recognition.

\vspace{-0.3 cm}
\subsubsection{\bf Deep learning methods}
Compared with the shallow machine learning methods that demand handcrafted descriptors to extract features, the deep learning for MiE recognition (a.k.a., deep MiE recognition) can directly learn the advanced semantic features from corresponding data and achieve state-of-the-art performance on MiE recognition.
Before proceeding, we first discuss the different types of input for MiE recognition. 
\vspace{-0.2 cm}
\begin{itemize}
\item \textbf{Static images and dynamic frames.} Given the subtle motions in per MiE, the apex with peak emotional intensity is often employed to reduce computational complexity \cite{zhou2019cross, gan2018bi, liong2018less}.
Li et al. \cite{li2018can} demonstrated the effectiveness of deep models using apex frames for MiE recognition.
Furthermore, Sun et al. \cite{sun2020dynamic} showed that apex frames performed better than complete video sequences in certain scenarios.
To capture dynamic information, several approaches utilize multiple sparse frames \cite{liu2020sma, gan2019off, xia2020revealing, zhu2023learning} or dynamic images \cite{nie2021geme,le2020dynamic,verma2019learnet,sun2020dynamic}.
Sparse frames allow for the extraction of richer information to describe MiE motion.
Liu et al. \cite{liu2020sma} proposed a dynamic segmented sparse imaging module to highlight key frames and describe subtle motion changes.
Gan et al. \cite{gan2019off} presented OFF-ApexNet to extract optical flow features from the onset to apex frames.
In addition, Xia et al. \cite{xia2020revealing} calculated optical flow maps from the onset to apex frames.
Zhu et al. \cite{zhu2023learning} proposed Learning to rank onset-occurring-offset features, which used onset, offset, and a random frame to reduce low-intensity information and enhance emotional expressiveness, achieving discriminative features.
\item \textbf{Temporal information.} Inspired by an action recognition method that converted dynamic clips into single dynamic images to integrate spatio-temporal information \cite{bilen2016dynamic}, multiple MiE recognition approaches \cite{nie2021geme,le2020dynamic,verma2019learnet,sun2020dynamic} use dynamic images to capture subtle movements during expression clips. 
Due to the incorporation of richer temporal information without incurring significant computational overhead, the methods using dynamic images can achieve better performance than apex frame-based approaches in MiE recognition.
\item \textbf{Complete clips.} Advances in deep models for time series processing allow extraction of spatio-temporal information from continuous clips and several methods demonstrate competitive performance \cite{li2019micro,zhao2021two,reddy2019spontaneous,bai2020investigating}. 
However, due to information redundancy and the limited scale of MiE datasets, such deep models may lead to high computational costs and overfitting.
\end{itemize}
\vspace{-0.2 cm}

Different deep learning paradigms have been explored to address various challenges in MiE recognition \cite{fan2023selfme, nguyen2023micron, das2021interpretable}. 
Therefore, we are motivated to discuss various learning paradigms for deep MiE recognition.

\textbf{Transfer learning.} Due to the scarcity of MiE datasets and the subtle intensity of MiEs, transfer learning for deep MiE recognition needs a stronger reliance on prior knowledge on facial expressions and the ability to capture subtle motion changes. Transfer learning in MiE recognition tends to learn features from MaEs and AUs.
For instance, Patel et al. \cite{patel2016selective} pre-trained models on both ImageNet and MaE datasets and demonstrated that MaE datasets are more suitable for MiE recognition due to their domain-specific features.
The application of transfer learning to deep MiE recognition involves two steps: (1) pre-training on related datasets (e.g., facial expressions and ImageNet datasets) to learn general features of human faces and objects; and (2) fine-tuning on MiE datasets. 
Moreover, transfer learning has been investigated by leveraging knowledge from MaE datasets that contain a large number of labeled \maes~\cite{peng2018macro,zhi2019combining,wang2020micro,xia2020learning,xia2021micro}.
For example, Zhi et al. \cite{jia2018macro} proposed to pre-train a 3D-CNN on the Oulu-CASIA dataset to enrich MiE features and improve recognition performance. 
Before fine-tuning on MiE datasets, Wang et al. \cite{wang2020micro} proposed to address the overfitting issue by sequentially pre-training network parameters on ImageNet and MaE datasets.
To better utilize dynamic muscle motion, Xia et al. \cite{xia2021micro} proposed a dual-stream deep model with one stream pretrained on MiE data and the other on MaE data.
The domain discriminator and the relation classifier were designed to account for spatio-temporal information during the transfer process.
In particular, the domain discriminator module focuses on capturing the static textures of facial appearances, and the relation classifier predicts the correct relation among temporal features with different sampling intervals from the double streams of the model. 
Indolia et al. \cite{indolia2022integration} revealed that fine-tuning a pre-trained ResNet18 model with a self-attention mechanism can identify relevant facial regions for MiE recognition.
Liu et al.\cite{liu2022lightweight} used MaE samples to train the feature extractor and optimized the hyper-parameters through grid search for subsequent deep MiE recognition.

As an alternative of fine-tuning, knowledge distillation involves transferring knowledge from a complex ``teacher'' model to a simpler ``student'' model. 
For example, Sun et al. \cite{sun2020dynamic} used a pre-trained teacher model to distill AU-based knowledge to guide the student model. 
Li et al. \cite{li2021micro} developed a dual-view attentive similarity-preserving distillation approach to learn AU knowledge and employed a semi-supervised co-training method to generalize the teacher model. 
Song et al. \cite{song2022recognizing} incorporated related expression samples into a first-order motion model to extract motion variations from neutral expressions to MaE and MiE. 
More specifically, Song et al. \cite{song2022recognizing} used a dual-channel encoder-decoder model guided by teacher model features to learn the transition features from MiE to MaE.

\textbf{Multi-task learning.}
Multi-task learning in deep MiE recognition focuses on incorporating fine-grained auxiliary tasks to enhance the model's sensitivity to subtle facial changes. Specifically, multi-task learning leverages auxiliary tasks (e.g., gender detection \cite{li2019facial, nie2021geme} and AU detection \cite{fu2022micro, wang2023action}) to improve model generalization and robustness in deep MiE recognition.
Li et al. \cite{li2019facial} utilized multiple related tasks (e.g., facial landmark detection, gender detection, smiling recognition, profile analysis, and glasses detection) as auxiliary tasks for MiE recognition. Similarly, Nie et al. \cite{nie2021geme} proposed a dual-stream approach based on gender for MiE recognition, demonstrating that integrating gender features can enhance MiE recognition performance.
As an auxiliary task in MiE recognition, Fu et al. \cite{fu2022micro} employed AU detection to extract fine-grained features, thereby improving the capabilities of attention-based models. 
Wang et al. \cite{wang2023action} incorporated AU detection to introduce inductive biases that enhance the generalization of deep MiE recognition.

\textbf{Self-supervised learning.}
To enhance the ability to capture details of MiEs, self-supervised learning can reduce noise and irrelevant information in subtle changes of MiEs. 
Fan et al. \cite{fan2023selfme} proposed a MiE analysis method by using self-supervised learning and a symmetric contrastive ViT to capture the symmetrical facial movements in MiEs while mitigating the influence of irrelevant information.
Since standard BERT cannot describe MiE features in detail, Nguyen et al. \cite{nguyen2023micron} introduced a self-supervised learning method called micron-BERT that consists of two components: diagonal micro-attention to capture subtle differences between adjacent frames and a patch-of-interest mechanism to emphasize regions within MiEs while mitigating background noise.
Das et al. \cite{das2021interpretable} elaborated a self-supervised approach to learning meaningful dynamic features of MiEs with limited samples as pretext tasks. 
The pretext model was subsequently adapted for downstream tasks while retaining prior knowledge of facial motion.
Motivated by contrastive learning within the momentum contrast framework \cite{he2020momentum}, Wang et al. \cite{wang2023temporal} utilized a pre-trained 3D-CNN to learn discriminative information embedded in the underlying data structure. To capture subtle and rapid facial movements, the 3D-CNN enhanced temporal dynamics by incorporating a generative adversarial model to remove redundant frames.

\textbf{Lightweight deep learning.}
The ever-increasing demand for MiE recognition in daily life drives the development of lightweight and resource-efficient deep MiE recognition methods \cite{khor2019dual, liong2019shallow, ab2023lightweight}.
Compared with the complex deep models, the lightweight deep model can alleviate the computational demand while maintaining reasonable predictive performance.
Khor et al. \cite{khor2019dual} proposed a lightweight dual-stream lightweight deep model comprising two truncated CNNs with distinct input features. 
By merging different convolutional features from both streams, the proposed method facilitates more effective training and enhances the learning of discriminative features for MiE recognition.
Liong et al. \cite{liong2019shallow} designed a triple stream 3D-CNN to recognize MiE from three optical flow features, i.e., optical strain, horizontal and vertical optical flow fields.
With fewer model parameters than traditional deep models, the proposed method in \cite{liong2019shallow}  could extract discriminative high-level features to describe the fine-grained MiE details.
Shukla et al. \cite{shukla2022micro} introduced a hybrid method combining convolutional layers with LSTM for lightweight MiE recognition. 
By integrating convolutional and recurrent structures, the proposed model \cite{shukla2022micro} can capture complex spatio-temporal features and address data imbalance issues. 
To identify global features, Wang et al. \cite{wang2023shallow} proposed a multi-branch attention CNN that consists of region division, multi-branch attention expression learning, and global feature fusion. 
The proposed multi-branch attention module can extract four distinct features that are combined with different weights.

\vspace{-0.4 cm}
\subsubsection{\bf Other promising learning paradigms.}
To expand the scope of MiE recognition, several promising techniques \cite{dai2021cross, li2021improved, sharma2022evaluation} have emerged that cater to practical requirements and specific application scenarios. 
These methods address challenges such as limited datasets and focus on applications across diverse settings.
One approach to train deep models with limited data is to employ meta learning. 
By training on a variety of tasks, meta-learning enables models to ``learn how to learn'', acquiring rapid adaptation capabilities that are valuable in data-scarce scenarios.
Wan et al. \cite{wan2022micro} applied model-agnostic meta learning to initialize parameters for 3D CNN training, enabling the model to quickly adapt to new tasks with minimal fine-tuning. 
Gong et al. \cite{gong2023meta} utilized a meta-learning-based multi-model fusion framework to extract the deep features from frame differences and optical flows. 
To address the insufficient and inconsistent labeling in MiE datasets, Dai et al. \cite{dai2021cross} introduced a few-shot learning method that transfers knowledge through AUs. 
The few-shot learning method in \cite{dai2021cross} leverages the rich information contained in AUs to enhance the model ability to generalize from limited labeled examples.
After verification of various feature combinations, the framework demonstrated that the fusion method using MaE and MiE can give distinct information to recognize MiEs.

	Another approach to facilitate insufficient samples in MiE recognition is data generation using a generative adversarial network (GAN) \cite{goodfellow2014generative}. 
	The proposal of GAN can be applied in many aspects including image generation \cite{liao2022text}, data augmentation \cite{tran2021data}, super-resolution \cite{park2023content}, and style transfer \cite{xu2021drb}. 
	Recently, MiE studies employed the GAN to generate more available data samples for training robust recognition models. 
	Liong et al. \cite{liong2020evaluation} used two GAN-based models, namely auxiliary classifier GAN  \cite{odena2017conditional} and self-attention GAN \cite{zhang2019self}, to generate artificial MiE samples. 
	Li et al. \cite{li2021improved} proposed an improved GAN for MiE analysis using AU-based multi-label learning to generate sequences that were very similar to the original sequences.
	This approach ensures high fidelity in the generated data, which is crucial for training robust MiE recognition models.
	To address the issues of AU annotation caused by a subtle change in the face, Xu et al. \cite{xu2021famgan} introduced a fine-grained AUs modulation to alleviate the noises and handle the symmetry of AUs intensity.
	Yu et al. \cite{yu2021ice} provided a GAN-based method, identity-aware and capsule-enhanced generative adversarial network with graph-based reasoning, to conduct controllable MiE synthesis with identity-aware features for recognition.  
	This innovative approach allows for the generation of synthetic MiE sequences that retain subject identity while enhancing recognition performance.
	Zhou et al. \cite{zhou2023ulme} developed a GAN-based approach to enhance the generation of MiE sequences, aiming for more natural and seamless appearances. 
	They first analyzed all AUs presented in prominent MiE datasets. 
	Then, they smoothed the AU matrix extracted from source videos to refine the input data. 
	By leveraging this refined AU information within their GAN framework, they were able to generate MiE sequences that exhibit more realistic and natural transitions.
	Zhang et al. \cite{10375342} introduced a facial-prior-guided MiE generation framework for facial motion synthesis. This framework incorporated two key structures: an adaptive weighted prior map and a facial prior module. 
	These components mitigated AU estimation errors and guided motion feature extraction, ensuring smooth and realistic synthesis generation by leveraging facial priors.
	
	The focus of research on MiE recognition has been shifting from controlled laboratory settings to uncontrolled, wild scenarios to better address practical applications.
	Accurate MiE recognition in such environments presents several challenges, including occlusions, pose variations, varying illuminations, and low-resolution images. 
	To address the challenges and improve the robustness and performance of recognition systems, a variety of techniques have been proposed.
	One approach to handling low-resolution MiE images in uncontrolled environments is through super-resolution techniques.
	Sharma et al. \cite{sharma2022evaluation,sharma2022comparative} introduced methods based on GAN to perform super-resolution on low-resolution MiE images. 
	This approach transformed low-resolution MiE images to super-resolution, allowing for the extraction of fine-grained textural features useful for downstream tasks. 
	Occlusions represent another significant challenge for MiE recognition in wild conditions. 
	Mao et al. \cite{mao2022objective} alleviated occlusions by generating synthetic datasets featuring various types of occlusions, such as facial masks, glasses, and random region masks, which allow for more thorough analysis under occlusion conditions.
	Gan et al. \cite{gan2022needle} developed an integrated ``spot-and-recognize'' framework that incorporated modules for MiE spotting, face alignment, and feature extraction. 
Building on this foundational work, Gan et al. \cite{gan2023revealing} introduced significant enhancements by integrating 3D face reconstruction techniques into the existing framework.
Specifically, Gan et al. \cite{gan2023revealing} developed methods to generate a 3D face mesh that improves the accuracy of face alignment and the reliability of feature extraction from a 3D perspective.
Even under challenging conditions such as varying poses and illuminations, the proposed integration in \cite{gan2023revealing}  can obtain accurate and consistent MiE recognition.

\vspace{-0.2 cm}
\subsubsection{\bf Integrated Comparisons \& Practical Insights on MiE Recognition}
\vspace{-0.2 cm}
Once a MiE is spotted, recognition becomes the next pivotal task: classifying the expression’s emotional content. Unlike MaE recognition, MiE recognition requires the system to decode extremely subtle cues that lack strong semantic or muscle activation signals. Traditionally, shallow machine learning methods have been used with handcrafted descriptors such as LBP/LBP variants \cite{wang2015lbp,wang2015efficient,wei2022micro}, HOG \cite{li2017towards, niu2018micro, zhang2020new}, and optical flow-based methods \cite{liong2014optical, happy2017fuzzy,happy2018recognizing,patel2015spatiotemporal}, offering interpretable yet limited representations. The emergence of deep learning revolutionized this space by enabling feature hierarchies to be learned from data, especially when combined with transfer learning \cite{patel2016selective,peng2018macro,jia2018macro,zhi2019combining,li2020joint,liu2020sma,chen2020spatiotemporal,wang2020micro,xia2020learning,sun2020dynamic,xia2021micro,li2021micro,indolia2022integration,liu2022lightweight,song2022recognizing}, multi-task learning \cite{patel2016selective,peng2018macro,jia2018macro,zhi2019combining,li2020joint,liu2020sma,chen2020spatiotemporal,wang2020micro,xia2020learning,xia2021micro,indolia2022integration,liu2022lightweight,sun2020dynamic,li2021micro,song2022recognizing}
, self-supervised learning \cite{fan2023selfme,nguyen2023micron,das2021interpretable,he2020momentum,wang2023temporal}
, lightweight deep learning \cite{khor2019dual,liong2019shallow,ab2023lightweight,shukla2022micro,wang2023shallow}
 and other learning paradigms \cite{dai2021cross,li2021improved,sharma2022evaluation,wan2022micro,gong2023meta,goodfellow2014generative,liao2022text,tran2021data,park2023content,xu2021drb,liong2020evaluation,odena2017conditional,zhang2019self,xu2021famgan,yu2021ice,zhou2023ulme,10375342,sharma2022comparative,mao2022objective,gan2022needle,gan2023revealing}
. 
A detailed comparison of deep MiE recognition methods is provided in Appendix 4.2.

Shallow learning offers a pragmatic advantage for low-data scenarios. For instance, descriptors (e.g., LBP, HOG, and optical flow) provide strong baseline performance with minimal computational overhead. However, these handcrafted features require domain-specific tuning and fail to generalize well across databases or spontaneous MiEs.
Deep learning methods address these limitations through end-to-end spatio-temporal modeling. 
However, their success hinges on rich data. Transfer learning has become an effective solution: models pre-trained on macro-expressions are fine-tuned on MiEs to leverage transferable visual features \cite{peng2018macro,xia2021micro}. 
Techniques such as knowledge distillation \cite{sun2020dynamic,song2022recognizing} have helped reduce model complexity and overfitting, while multi-task learning (e.g., with AU or gender detection \cite{nie2021geme,wang2023action}) boosts sensitivity to subtle cues by introducing regularization via auxiliary tasks.
Meta-learning \cite{gong2023meta,dai2021cross} supports cross-domain generalization, enabling more robust recognition across datasets with different recording conditions (e.g., CASME II vs. SAMM). 
GAN-based technique provides a way to increase data samples by generating synthetic MiE images.
In addition, lightweight deep learning architectures offer a practical foundation for deploying MiE recognition models on edge devices. 
Experimental results from \cite{khor2019dual, liong2019shallow} indicate that models with fewer than 1 million parameters can still achieve competitive accuracy (e.g., 83.82\% on CASME II, 65.88\% on SAMM).
Self-supervised uses temporal information and motion continuity to improve MiE recognition capabilities, thereby increasing the accuracy of the model (92.9\% on CASME II \cite{wang2023temporal}).

\vspace{-0.4 cm}
\section{Potential Applications of Facial Expression Analysis in IoT}\label{appl}\vspace{-0.2 cm}
The convergence of facial expression analysis with IoT systems unlocks new chances for intelligent, pervasive and adaptive computing services. 
By interpreting human emotions, IoT-based applications can provide personalized, efficient, and responsive user experiences.
In this section, we explore the potential of facial expression analysis in IoT by separately examining the contributions of MaE recognition and holographic MiE analysis.

\vspace{-0.4 cm}
\subsection{MaE and Its Applications in IoT}\label{maeiot}\vspace{-0.2 cm}
MaEs play a vital role in interpreting human emotions; and therefore, MaE recognition becomes valuable in various IoT applications.
More specifically, the applications of MaE recognition based IoT systems have been extended from purely emotion recognition \cite{muhammad2021emotion,chen2020facial,yang2021real} to the healthcare industry \cite{hossain2019audio} and flu tracking \cite{rahman2021internet}.
	
\vspace{-0.3 cm}
\subsubsection{\bf Emotion recognition.} Recent years have witnessed the integration of emotion recognition with intelligent IoT devices for enhancing real-time performance, minimizing communication overhead, and protecting data privacy. 
Muhammad et al. \cite{muhammad2021emotion} proposed an edge-computing-based system that performs pre-processing on IoT devices and inference on edge servers. 
Pre-processing tasks (e.g., face detection, cropping, contrast enhancement, and resizing) can be performed on IoT devices to alleviate computation and communication expenditure. 
Then, the edge device leverages the processed images for inference and decision-making and returns the decisions to the IoT devices.
Moreover, the edge device downloads a global deep model from the cloud during off-peak hours to minimize latency.
Chen et al. \cite{chen2020facial} ensured privacy by processing sensitive emotional data locally. 
Yang et al. \cite{yang2021real} developed a real-time system using lightweight models and minimal transmission. 
Chen et al. \cite{chen2022emotion} applied emotion detection in vehicles, achieving 2 ms latency and a 96\% F1 score through optimized IoT protocols.

\vspace{-0.3 cm}
\subsubsection{\bf Healthcare industry.} 
The application of MaE analysis in the IoT-based healthcare industry has demonstrated a promising potential. 
By integrating the MaE information with IoT technology, healthcare systems can provide real-time analysis and feedback that are crucial for improving medical services and patient care.
For example, Hossain et al. \cite{hossain2019audio} developed an emotion recognition system for medical facility enhancement. 
The proposed approach utilized an IoT-based audio-video framework to assist healthcare providers in evaluating services. 
The system performs audio feature extraction and face detection on edge devices.
Then, the remote cloud delivers the well-trained parameters to edge devices for efficient model inference. 
Such setup allows healthcare providers to gain immediate insights into patient emotions, facilitating more personalized and responsive care. 
Wang et al. \cite{wang2022edge} highlights the potential of IoT-based MaE analysis in ensuring worker safety in hazardous industries.
More specifically, a deep capsule method based on IoT devices was introduced in \cite{wang2022edge} to monitor the mental state of miners in underground mining environments. 
The proposed system collects facial expressions and electroencephalograms to infer mental health states. 
By integrating the two modalities, the system could detect signs of stress and fatigue and thereby reduces the risk of man-made accidents.

\vspace{-0.3 cm}
\subsubsection{\bf Flu tracking.}
The application of IoT-based MaE analysis in flu tracking enhances pandemic management through real-time monitoring and data analysis. 
For instance, Rahman et al. \cite{rahman2021internet} developed an edge Internet of Medical Things framework to capture signals (e.g., facial expressions and psychological states) during pandemics. 
By leveraging the power of edge computing, the framework processes data locally to generate immediate reports with key insights. 
The local processing capability ensures low latency, privacy preservation, and enhanced security that are critical for managing sensitive health data. 
Additionally, the framework supports various applications, e.g., sleep analysis, face mask detection, and physiological state monitoring during pandemics.

\vspace{-0.4 cm}
\subsection{MiE and Its Applications in IoT}\label{mieiot}\vspace{-0.2 cm}
Since MiE can reveal ground-truth emotions, MiE-based IoT applications are an emerging and promising application scenario. 
By incorporating MiE analysis, IoT-based frameworks can detect and identify hidden emotions such that the valuable insights for practical applications are provided, e.g., security monitoring \cite{WOS:000615785700014,yildirim2023influence,chung2014empirical,shilaskar2023expert}, interpersonal negotiation \cite{xiong2023smart}, and mental disorder detection \cite{WOS:000703007300023, das2021interpretable, WOS:000853232800019, WOS:001123458900001, chen2023catching}.

\vspace{-0.1 cm}
\subsubsection{\bf Security monitoring.}
\mies~ are novel social behavioral biometrics that can be utilized for security monitoring. 
Due to the natural characteristics, \mies~ provide valuable biometric information that is difficult to pose or fake. 
Saeed et al. \cite{WOS:000615785700014} explored the use of \mies~ as a soft biometric for person recognition. 
The person recognition utilized three handcrafted features to extract facial and behavioral biometric features from videos containing \mies~ and predicted potential attacks.
\mies~ can also be used for deception detection. 
For instance, Yildirim et al. \cite{yildirim2023influence} made substantial contributions by analyzing facial expressions in video data.
They assessed the likelihood of the most prominent MiEs within a given clip or image to achieve accurate deception detection. 
Widjaja et al. \cite{widjaja2023exploring} introduced a deception detection system based on the GRU. 
This system analyzed meaningful patterns from MiEs in facial videos to identify potential signs of deception. 
By leveraging the temporal dynamics captured by GRUs, the system can detect subtle changes in facial expressions over time, enhancing its accuracy in identifying deceptive behavior.
Shilaskar et al. \cite{shilaskar2023expert} also examined the effectiveness of deception detection using MiEs. 
Their proposed computer vision-based method evaluates \maes~ and \mies~ captured by a camera to identify specific facial cues associated with deception.

\vspace{-0.35 cm}
\subsubsection{\bf Interpersonal negotiation.}
Although emotion can be reflected in facial expressions, it is difficult for someone to capture the emotional changes in time due to the complexity of actual situations. 
To capture the true emotion and provide timely feedback in negotiation, Xiong et al. \cite{xiong2023smart} introduced a real-time MiE-based emotion recognition system via wearable smart glasses. 
By collecting and extracting features of the facial video on the front end of smart glasses, users can understand the communicator's intended information through real-time analysis of MiEs.

\vspace{-0.35 cm}
\subsubsection{\bf Mental disorder detection.}
As fleeting information in facial activity, the subtle movements of MiE can reveal psychological states. 
Huang et al. \cite{WOS:000703007300023} proposed an elderly depression detection method using the \mies~ based on AU features. 
Das et al. \cite{das2021interpretable} presented a method to extract significant MiE muscle movements to predict mental disorders and demonstrated its potential in detecting future stuttering disfluency. 
Gilanie et al. \cite{WOS:000853232800019} proposed a lightweight depression detection method using CNN for real-time applications via digital cameras. 
Li et al. \cite{WOS:001123458900001} achieved promising results in depression recognition by analyzing the \mies. 
Chen et al. \cite{chen2023catching} developed a method to diagnose concealed depression using MiE combined with ROI and machine learning, offering a low-cost, privacy-preserving solution feasible for self-diagnosis on mobile devices.

\vspace{-0.4 cm}
\section{Challenges and Future Directions}\label{sec7}\vspace{-0.1 cm}
Despite considerable progress in facial expression analysis, the path toward practical, real-world deployment remains fraught with challenges that include data-related limitations (e.g., small size, low inter-rater agreement), model-related issues (e.g., overfitting and low generalization), and system-level constraints (e.g., user privacy and real-time processing). 
More specifically, \textbf{facial expression datasets remain limited due to their transient nature and complex annotation requirements \cite{muhammad2021emotion}.} Most MiE datasets are collected under controlled conditions, restricting real-world applicability. Future work should emphasize unsupervised learning, synthetic data generation using GANs or diffusion models, and standardized data collection protocols to improve scalability.
{\bfseries Cultural adaptation remains crucial.} While MiEs are considered cross-culturally consistent, MaEs vary widely due to cultural display rules \cite{yang2021real}. 
Regionally diverse datasets and contextual metadata (e.g., social setting and hierarchy) are necessary for improving model generalizability. Transfer learning and contextual fusion with environmental signals can enable systems to offer culturally appropriate responses.
{\bfseries data privacy is a major concern in IoT-based facial analysis.} Facial data, being biometric, poses security risks during collection and transmission via unsecured or wireless channels. Synthetic data generation (e.g., with Sora) and federated learning offer privacy-preserving alternatives, keeping data local while enabling collaborative training \cite{chen2020facial}. However, optimization is needed for deployment on resource-constrained devices.
{\bfseries Real-time processing is essential in applications like emotion-aware assistants or medical monitoring.} High latency from cloud processing must be avoided. Solutions include edge computing, model compression techniques (e.g., quantization, pruning, distillation), and event-driven models using neuromorphic cameras \cite{chen2022emotion}. Lightweight communication protocols are vital to sustain real-time performance in bandwidth-limited scenarios.
Finally, comprehensive analysis using multi-modal inputs such as electromyography \cite{kim2022classification}, electroencephalogram \cite{wang2023micro, zhao2024micro}, and psychological cues \cite{saffaryazdi2022using} can enhance emotion recognition. Efficient multi-task learning models are also needed to manage diverse sensor inputs in complex IoT environments \cite{chen2019analyze, zou2022concordance}. These efforts will support human-centric, context-aware services in future IoT systems. Comprehensive discussions can be found in Appendix 5.

 \vspace{-0.2 cm}
 \section{Conclusions}\label{sec8}\vspace{-0.2 cm}
 Facial expression analysis is receiving perpetually rising attention to understand the rich emotional information. 
 As a common type of facial expression, the MaE analysis is now used in various wild scenarios.
 In contrast, the MiE analysis with the characteristics of unconsciousness and subtlety has the potential to reveal the genuine emotions of each individual.
 In this article, we have provided the taxonomy of current facial expression analysis that includes MaE and MiE analysis.
 We have comprehensively reviewed the state-of-the-art facial expression analysis methods and discussed their corresponding limitations. 
 By reviewing the current applications of MaE and MiE in IoT systems, we have discussed insights and potential directions for the development and implementation of facial expression analysis in future IoT systems. 
 We expect that our work can serve as a valuable resource for researchers and practitioners in facial expression analysis by providing fundamental research resources and a comprehensive research paradigm. 

\vspace{-0.2 cm}
\bibliographystyle{abbrvnat}\vspace{-0.2 cm}
\bibliography{ref2}

\section{Nomenclature}
	Here is the nomenclature of our work.

	\begin{tabbing}
		\textbf{Abbreviations} \:\= \textbf{Definitions} \\
		AFEW \> The acted facial expressions in the wild \\
		AU \> Action unit \\
		BERT \> Bidirectional encoder representations from transformers \\
		BU-3DFE \> Binghamton university 3D facial expression \\
		CAER \> Context-aware emotion recognition \\
		CASME \> Chinese Academy of Sciences micro-expression \\
		CASME II \> Chinese Academy of Sciences micro-expression II \\
		CAS(ME)$^2$ \> Chinese Academy of Sciences  macro-expressions and micro-expressions \\
		CK+ \> The Extended Cohn-Kanade Dataset \\
		CNN \> Convolutional neural network \\
		DFEW \> Dynamic facial expression in the wild \\
		ExpW \> Expression in-the-wild \\
		FACS \> Facial action coding system \\
		FER-2013 \> Facial expression recognition 2013 \\
		GAN \> Generative adversarial network \\
		HOG \> Histogram of gradient \\
		ISED \> Indian spontaneous expression database \\
		IoT \> Internet of Things \\
		JAFFE \> Japanese female facial expression \\
		KDEF \> Karolinska directed emotional faces \\
		LBP \> Local binary pattern \\
		LBP-TOP \> Local binary pattern from three orthogonal planes \\
		LSTM \> Long short-term memory \\
		MaE \> Macro-expression \\
		MDMO \> Main directional mean optical-flow \\
		MEVIEW \> Micro-expression videos in the wild \\
		MiE \> Micro-expression \\
		MMEW \> Micro-and-macro in expression warehouse \\
		PEDFE \> Padova emotional dataset of facial expressions \\
		RaFD \> Radboud faces database \\
		RNN \> Recurrent neural network \\
		ROI \> Region of interest \\
		SAMM \> Spontaneous actions and micro-movements \\
		SFEW 2.0 \> Static facial expressions in the wild \\
		SMIC \> Spontaneous micro-expression corpus \\
		TFD \> The Toronto face database \\
		UAR \> Unweighted average recall\\
		ViT \> Vision transformer \\
		WAR \> Weighted average recall 
	\end{tabbing}

	\section{Illustrations of Edge-Driven Facial Expression Analysis Systems, MiEs, and MaEs}
	\setcounter{figure}{1}
	\begin{figure}[htb]
		\vspace{-0.5 cm}
		\begin{minipage}{0.5\textwidth}
			\centering
			\includegraphics[width=6.6 cm, height=5.5 cm]{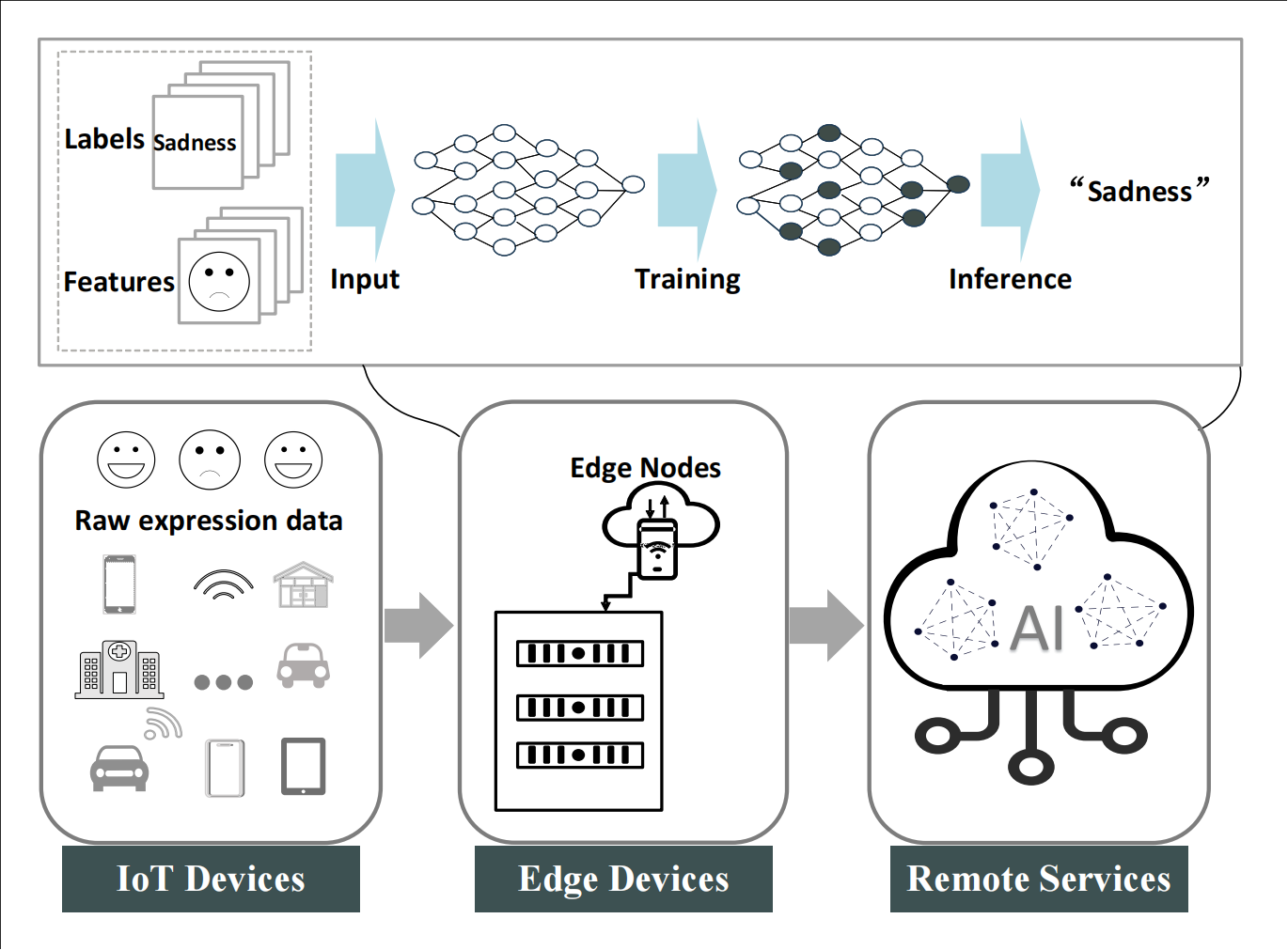}
			\captionof{figure}{A framework of edge-driven facial expression analysis.}
			\label{fig1}
		\end{minipage}
		\hspace{0.5 cm}
		\begin{minipage}{0.43\textwidth}
			\centering
			\vspace{0.35 cm}
			\includegraphics[width=0.85\textwidth, height=5.5 cm]{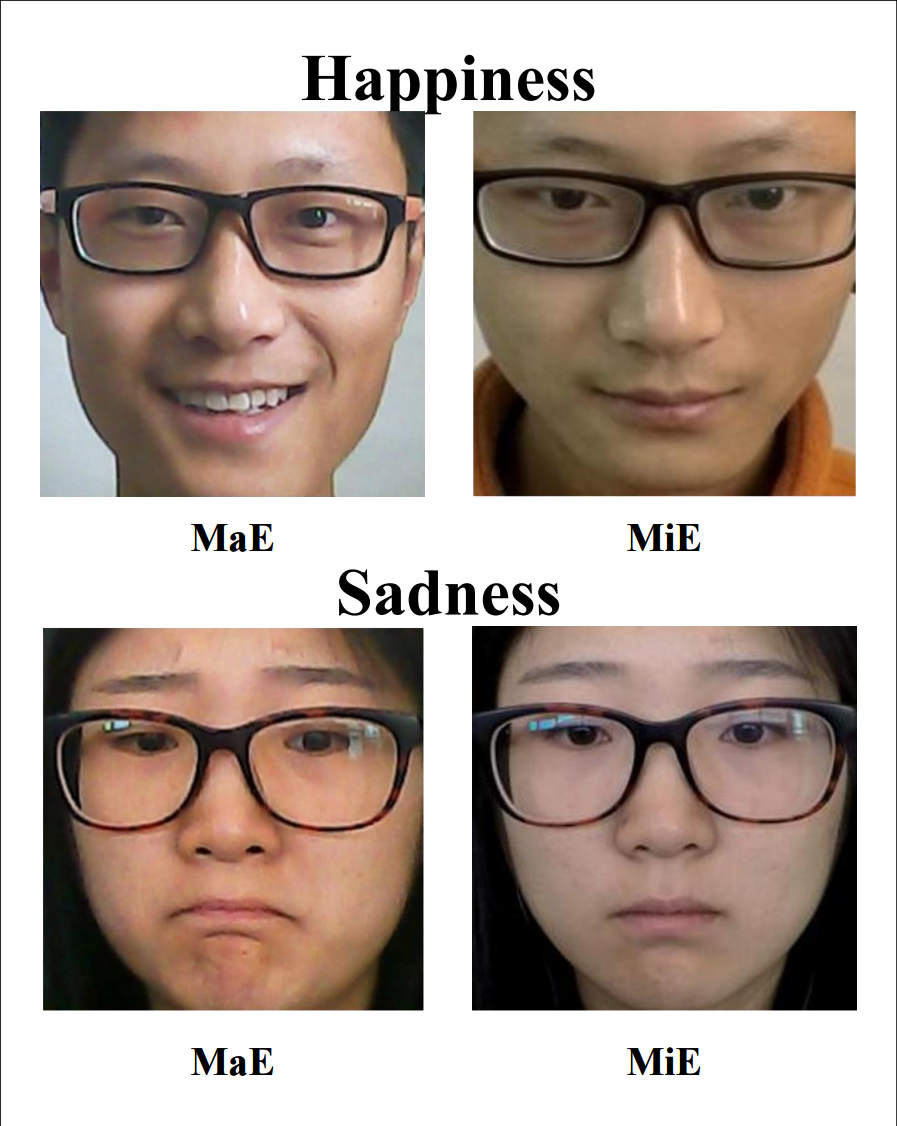}
			\captionof{figure}{An illustration of \maes~and~\mies~ in micro-and-macro in expression warehouse.}
			\label{fig2}
		\end{minipage}%
		\vspace{-0.5 cm}
	\end{figure}
	
	As shown in Fig. \ref{fig1}, the IoT devices can first collect raw facial expression data for pre-processing. 
	To fully leverage low access latency and high data security during the implementation of real-time privacy-preserving systems, the pre-processed features are then uploaded to edge devices for decision-making. 
	
	Facial expressions can be divided into \maes~ and \mies~ based on their duration and intensity. 
	In terms of duration, \maes~approximately last 0.5 to 4 seconds, while \mies~last less than 0.5 seconds \cite{shen2012effects}.
	The short duration of \mies~conveys subtle emotional message, which makes emotion perception from \mies~a challenging task without specialized techniques and training. 
	As shown in Fig. \ref{fig2}, the major differences between the MaEs and MiEs are as follows.
	\begin{itemize}
		\item 	\textbf{\maes:} can be perceived during regular interactions. 
		The recognition accuracy of \maes~ can exceed 97\% in a controlled laboratory due to the spontaneity and noise immunity. 
		\maes~recognition can be applied in multiple scenarios, e.g., autonomous driving, psychological health assessment, and human-computer interaction; 
		\item  \textbf{\mies:} are involuntary and rapid. 
		Due to the short duration and subtle emotional cues, the detection of \mies~ can be challenging with naked eyes. 
		It is reported that the average recognition accuracy of~\mies~is around 47\% after specialized training \cite{frank2009see}. 
		Nevertheless, MiEs can reveal the true emotions of a person since they are instinctive and cannot be concealed. 
		Numerous MiE applications in IoT systems are being explored, such as lie detection, criminal identification, and security control.
	\end{itemize}
	
	\section{Structure of Our Work}
\begin{figure}[htp]
	\centerline{\includegraphics[width=\linewidth]{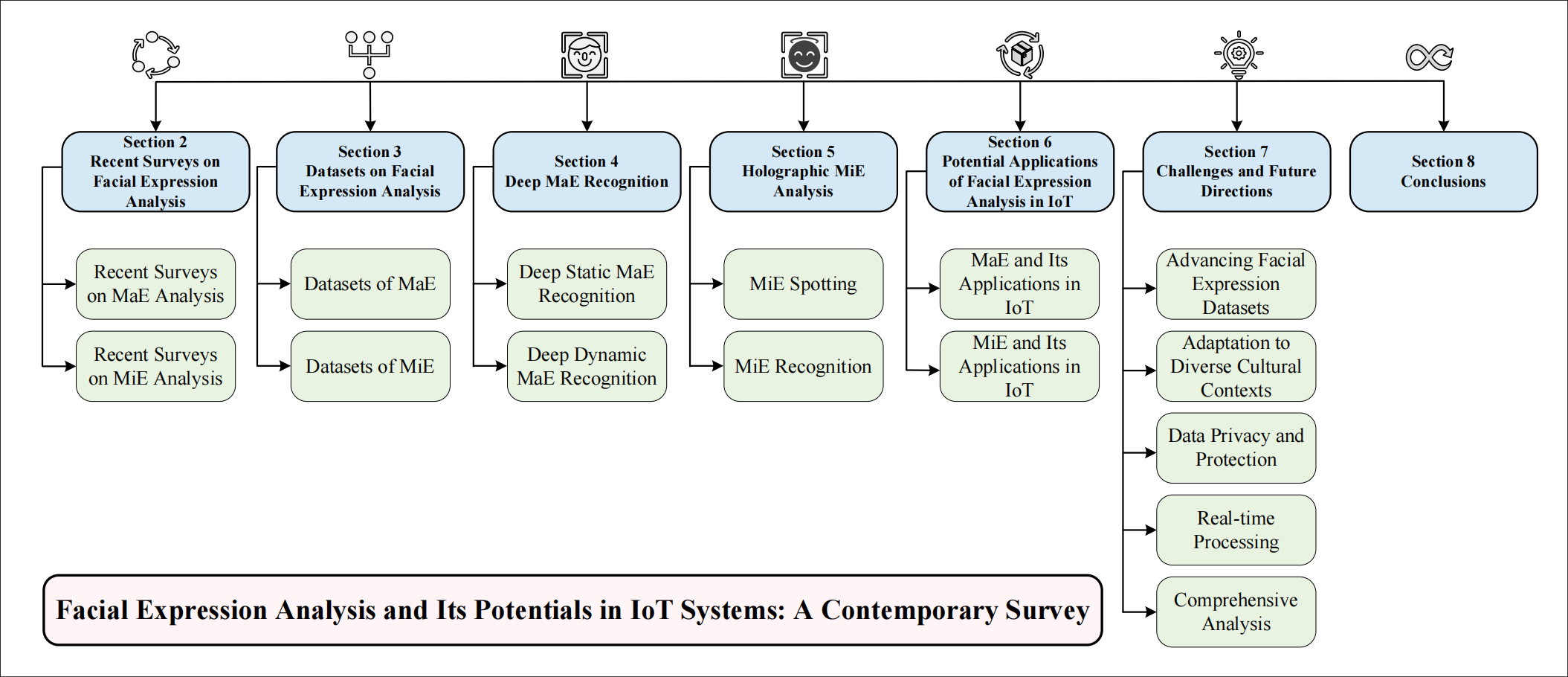}}
	\caption{The organization of the remaining work.}
	\label{supp:structure}
\end{figure}

	The remaining work is organized as follows as shown in Fig. \ref{supp:structure}. 
	Section 2 introduces the recent surveys on MaE and MiE analysis and the major differences from our work.
	Section 3 presents the datasets for MaE and MiE analysis. 
	Section 4 discusses the deep MaE recognition.
	Section 5 discusses the holographic MiE recognition that includes spotting and recognition procedures. 
	Section 6 discusses the applications of MaE and MiE analysis in IoT systems.
	Section 7 discusses the challenges and future directions. 
	The conclusions are presented in Section 8. 

\section{Comparison of Learning Paradigms for MaE Analysis}
	\subsection{Comparison of Learning Paradigms in Deep Static MaE recognition} 
	\textcolor{black}{
		Deep static MaE recognition focuses on analyzing facial expressions from multiple individual images by leveraging advanced learning paradigms to extract effective features that contribute to improving model accuracy. For example, transfer learning initially uses ImageNet for pretraining to learn generic object representations for MaE recognition \cite{10.1145/2818346.2830593,electronics10091036}.
	To better model facial geometry and feature distributions, researchers introduced MaE‑specific datasets, such as face‑identification data \cite{s20092639} and high‑quality MaE datasets \cite{Atabansi_2021}, for pretraining.
	However, domain shift remains a significant challenge, particularly when adapting models trained on controlled datasets to in-the-wild conditions (e.g., accuracy on AffectNet and the Emotion Recognition in the Wild Challenge is 60.70\% \cite{s20092639} and 55.6\% \cite{10.1145/2818346.2830593}).
	To mitigate this gap, multi‑task learning jointly optimizes MaE recognition with related objectives.
	Using auxiliary tasks such as AU detection and facial landmark localization improves recognition to 65.80\% on AffectNet and 98.33\% on CK+ \cite{pons2018multi,CHEN2021107893,YU2022108401,10214365}.
	Further gains have been achieved by incorporating pose estimation \cite{chen2022orthogonal,zaghbani2022multi}, gender classification \cite{9945168}, expression synthesis \cite{9413000}, and other auxiliary tasks \cite{10216308,FOGGIA2023105651}, pushing CK+ accuracy to 99.08\%.
	However, multi‑task learning presupposes datasets that include annotations for every auxiliary task, yet such datasets are rare in the MaE domain.
	To reduce the cost of manual annotation, self‑supervised learning derives contextual facial cues directly from unlabeled data \cite{9511468,Wang_2022_ACCV,CHEN2023206,10121455,an2024self}.
	By means of designed pretext tasks that model spatial relationships and semantic dependencies within a single image, these methods produce facial representations that transfer effectively to downstream MaE recognition.
	Moreover, self‑supervised learning can fuse multi‑view images \cite{roy2021self,10.1145/3632960} and cross‑modal data \cite{9206016,electronics12020288, halawa2024multi} to learn richer and more comprehensive representations for MaE recognition. 
	Self-supervised learning can achieve satisfactory results (e.g., accuracy on AffectNet and the CK+ is 66.04\%, 98.77\%), but their pretext tasks are computationally expensive, typically demanding large batch sizes, more training iterations, and sophisticated optimization strategies.
}

Ensemble learning combines heterogeneous features or models to exploit their complementary strengths. 
	The integration form of ensemble learning is scalable, supporting integration at several points in the pipeline—for example, input‑level fusion \cite{asd11,HARIRI202184,9226437,SHAO201982,VISWANATHAREDDY202023}, feature‑level aggregation \cite{8371638,9767587}, and decision‑level fusion \cite{9226437,8643924}.
	Drawing on multiple, diverse components yields more stable performance than any single model. 
	For instance, \cite{8371638} reports that ensembling the best single networks enhances accuracy from 95.16\% to 97.31\% on CK+ and from 92.36\% to 97.14\% on JAFFE.
	However, ensemble learning requires the integration of multiple models or pipelines, thus multiplying the model's resources in the training and inference stages.
	In addition, in static MaE recognition, ensemble schemes that operate above the feature level aggregate the outputs of independent components instead of learning a richer complementary representation, thereby limiting their ability to generalize.
	Attention mechanisms enrich MaE representations by spotlighting salient facial cues and modeling local–global dependencies.
	CNN‑based attention modules have raised recognition accuracy to 64.54\% on AffectNet and 98.68\% on CK+ \cite{itti2001computational,Fernandez_2019_CVPR_Workshops,LI2020340}, while simultaneously alleviating pose variation and occlusion through region‑focused weighting \cite{8576656,wang2020region,9474949,9750079}. 
	Transformer architectures push this idea further: their global self‑attention captures long‑range interactions among facial patches, boosting accuracies to 67.44\% on AffectNet and 99.80\% on CK+.
	Nonetheless, attention‑based approaches—especially transformer‑based architectures—require substantial GPU memory and computational resources.

\subsection{Comparison of Learning Paradigms in Deep Dynamic MaE recognition}
Deep dynamic MaE recognition involves analyzing facial expressions from video sequences by capturing their temporal progression. 
In contrast to static approaches that process individual frames, dynamic methods leverage the temporal evolution of expressions to achieve more comprehensive and context-aware recognition.
Ensembling the probabilities of individual frames offers an intuitive approach to extending MaE analysis from a static to a dynamic setting \cite{kahou2013combining, kahou2016emonets}.
Although conceptually straightforward, this strategy discards temporal ordering and therefore under‑exploits the temporal information of MaE video; on the AFEW benchmark it saturates at roughly 48\% accuracy.
To further model dynamic patterns, explicit spatio-temporal learning leverages model architectures such as recurrent networks (e.g., RNNs, LSTMs \cite{7890464, 9102419, 9134869, 9330541, 10.3745/JIPS.01.0067, 9206081, YU201850, an2020facial, 9051332}) and 3D CNNs \cite{Hasani_2017_CVPR_Workshops, 8845493, 8373844} to extract spatial and temporal features from MaE sequences.
Empirical results show that spatio-temporal models significantly improve accuracy—e.g., from 85\% to 89.5\% on CK+ and from 55\% to 67\% on MMI \cite{Hasani_2017_CVPR_Workshops, 8845493}.
Furthermore, incorporating facial landmark information into spatio-temporal frameworks has been shown to yield even more promising results, achieving 98.5\% on CK+ and 81.18\% on MMI \cite{Hasani_2017_CVPR_Workshops, 8845493, 7890464}.
These findings suggest that dynamic MaE recognition can benefit from integrating multiple complementary feature modalities. 
For example, the feature-level ensemble learning \cite{8863974, LIU2022182, LIU2023109368, PAN2024120138} has demonstrated that the integration of short-term expression dynamics and long-range contextual dependencies can improve the model accuracy (e.g., 92.5\% on MMI, 65.85\% on DFEW).
In addition, multi-task learning approaches \cite{yu2020facial, Jin_2021_ICCV, XIE2023126649} enhance the representation capability of dynamic MaE models by incorporating complementary information from auxiliary tasks (e.g., AU detection, facial region analysis), thereby improving model accuracy (e.g., 98.77\% on CK+, 90.40\% on Oulu-CASIA).
Although incorporating richer features can improve model performance, the effectiveness of the model is still challenged under in-the-wild conditions such as occlusion, pose changes, and complex backgrounds.

To further model the dynamic features of MaEs both in the laboratory environment and in the wild, attention-based learning has been proposed to capture informative cues by introducing the attention modules that can capture informative cues from global context, local details, and channel-wise dependencies.
By focusing on salient spatial regions, expressive temporal segments, and discriminative feature channels for modeling dynamic MaE features, attention modules \cite{8803603, 9209166, LIU2020145, SUN2021378, xia2022multi} can improve the recognition accuracy in laboratory conditions (e.g., 99.69\% on CK+, 91.25\% on Oulu-CASIA) and in the wild conditions (59.79\% on AFEW, 63.71\% on AffectNet).
Building on this, transformer-based models enhance dynamic MaE recognition by modeling long-range spatial and temporal dependencies.
Through their global self-attention mechanism, transformers are capable of capturing the complex temporal evolution of facial expressions, particularly in unconstrained environments.
With the implement of transformer \cite{10.1145/3474085.3475292, HUANG202135,ma2022spatio,CHEN2024123635}, the performance of dynamic MaE recognition has been improved 
For example, on in-the-wild benchmark datasets such as DFEW, FERV39k, and AFEW, the transformer-based model \cite{CHEN2024123635} has achieved UAR and WAR scores of 58.65\%/69.91\%, 41.91\%/50.76\%, and 52.23\%/55.40\%, respectively.
In addition, the transformer also supports capturing high-level semantic features across various modalities (e.g., text, audio) for dynamic MaE recognition due to its encoding capability \cite{Zhang_2022_CVPR,10250883}.
However, all of the above methods follow the form of supervised learning, which is limited by the training samples of the current MaE datasets.
To address these limitations, self-supervised learning has emerged as a promising alternative by leveraging unlabeled data to learn transferable dynamic MaE representations.
Empirical results show that self-supervised learning can achieve state-of-the-art performance in dynamic MaE recognition with UAR and WAR scores of 66.85\%/43.12\%, 63.41\%/74.43\% on FERV39k and DFEW.
Nonetheless, self-supervised learning requires a carefully designed pretext task that can learn effective dynamic information, and the training process often involves high computational costs due to large batch sizes and complex optimization schedules.

\section{Comparison of Learning Paradigms for MiE Analysis}
\subsection{Comparison of Learning Paradigms in MiE Spotting} 
MiE spotting—the task of detecting the precise onset and offset frames of a MiE within a video—is a precursor to successful MiE recognition. It demands precise temporal sensitivity to localize transient facial cues often embedded within long, expression-neutral sequences. 
Early work relied on handcrafted motion descriptors. Optical‑strain‑based methods \cite{patel2015spatiotemporal,wang2016main,wang2017main} and texture operators such as LBP‑TOP \cite{pfister2011recognising,esmaeili2020automatic} constituted the first generation of MiE spotters.
Under controlled settings, Pfister et al. \cite{pfister2011recognising} detected apex frames with LBP‑TOP; Yan et al. \cite{yan2015quantifying} combined constrained local models with LBP. 
Although these approaches approached manual performance, their accuracies on CASME and CASME II were still modest ($\approx$ 40\%), and feature engineering demanded substantial expertise.
Subsequent studies sought to improve robustness through alternative domains and learned feature selection. 
Li et al.\cite{li2018can,li2020joint} observed that apex frames concentrate the most discriminative motion and, by analysing frequency spectra, raised accuracies to 60.82 \% (CASME) and 65.02\% (CASME II). 
Esmaeili et al. \cite{esmaeili2022spotting} employed a CNN to select LBP planes that maximally capture micro‑movements, boosting spotting rates to 93\% and 89\% on the same datasets.

Interval spotting extends the task from single peaks to full MiE sequences. Traditional pipelines follow a four‑stage procedure: feature extraction, feature‑difference analysis, thresholding, and peak search. 
Early evaluations counted a detection as correct when the spotted peak lay inside an enlarged reference window around the ground‑truth interval. 
Using this metric, Li et al.~\cite{li2017towards} compared LBP and HOOF, reporting area under the curve (AUCs) of 83.32\% (SMIC‑E‑HS) and 92.98\% (CASME II). 
Han et al.’s collaborative feature difference method \cite{han2018cfd}, which combines Fisher‑weighted LBP (texture) and MDMO (motion), further improved AUCs to 97.06\% and 94.19\%, respectively.

More recent benchmarks adopt intersection‑over‑union (IoU) to judge whether a predicted interval overlaps sufficiently (IoU $\geq$ 0.5) with the ground truth. 
Under this stricter regime, Pan et al.’s fine‑grained Local Bilinear CNN \cite{pan2020local} obtained F1 scores of 5.95\% on CAS(ME)\textsuperscript{2} and 8.13\% on SAMM‑LV. 
Yin et al.’s AU‑aware GCN \cite{yin2023aware}, which embeds prior AU statistics into a graph‑based backbone and couples it with temporal interaction, advanced the state of the art to 38.34\% and 37.28\% on the same datasets.

\subsection{Comparison of Learning Paradigms in MiE Recognition} 
Shallow machine-learning approaches to MiE recognition are dominated by hand-crafted descriptors, which can be grouped into three families: (i) LBP-based, (ii) gradient-based, and (iii) optical-flow–based descriptors.
LBP and its variants remain the de-facto choice for modelling fine-grained texture changes produced by MiE.
Wang et al. \cite{wang2015lbp,wang2015efficient} proposed several LBP-TOP variants that reduce feature redundancy and pushed recognition to 45.75\% on CASME II and 55.49\% on SMIC. 
Wei et al. \cite{wei2022micro}  incorporated oblique motion modelling into LBP-TOP, raising accuracy to 70.00\% on CASME II and 67.86\% on SMIC-HS. 
However, LBP-based methods enable a substantial growth in dimensionality due to the sliding-window stacking required to retain motion cues.
Because optical flow explicitly characterises pixel-level motion (direction and speed), it is a natural proxy for the minute deformations that constitute a MiE. Liong et al. \cite{liong2014optical} were the first to exploit optical-flow strain, reaching 53.56\% on SMIC. 
Happy et al. \cite{happy2018recognizing} further encoded local motion via an angular histogram of optical-flow directions, improving performance to 67.21\% on CASME and 56.78\% on CASME II. These methods, nevertheless, incur high computational costs and are sensitive to head movement and illumination changes.
Gradient descriptors encode local orientation information and are more robust to lighting variation than LBP. 
Li et al. \cite{li2017towards} showed that gradient-based descriptors outperform LBP on colour videos, achieving 68.29\% on SMIC-HS and 67.21\% on CASME II.
Zhang et al. \cite{zhang2020new} partitioned the face into seven regions and extracted regional gradient features, yielding 53.04\% on CASME II and 47.56\% on SMIC. 
Since MiE's motion is very subtle, the features extracted by the gradient-based descriptor may lose fine-grained details.

Transfer learning has emerged as a practical remedy for the chronic scarcity of MiE data and the inherently subtle nature of MiE signals.
Peng et al. \cite{peng2018macro} first demonstrated that features learned from MaE datasets such as CK+ and Oulu-CASIA can be repurposed for MiE recognition, boosting accuracy to 70.6\% on SAMM and 75.7\% on CASME II.
Building on this idea, Xia et al. \cite{xia2021micro} introduced a two-stream CNN that is jointly pre-trained on MaE and MiE sequences, allowing spatial and temporal cues to be shared; their model attains 76.4\% on SAMM and 79.9\% on CASME II.
These studies confirm that a conventional ``pre-train-then-fine-tune'' pipeline offers a convenient performance lift by providing pertinent MaE representations. Nonetheless, the large amplitude of MaE can mask the fine-grained cues that characterize MiE.
To attenuate this mismatch, recent work has adopted knowledge-distillation paradigms that steer the learner toward the most informative facial regions or motion patterns.
Sun et al. \cite{sun2020dynamic} distill AU knowledge into the MiE branch, surpassing fine-tuning baselines with 86.7\% on SAMM and 72.6\% on CASME II (vs. 84.1\% and 66.6\%, respectively).
Complementarily, Song et al. \cite{song2022recognizing} distill a teacher that magnifies expression intensity, raising UAR to 75.5\% on SAMM and 93.7\% on CASME II.
While distillation mitigates the over-reliance on irrelevant MaE cues, it does increase training complexity and its efficacy remains sensitive to the particular MaE task chosen for the teacher.

Sharing representations across related tasks has been investigated as a way to mitigate the chronic data scarcity of the MiE dataset.
Nie et al. \cite{nie2021geme}  added a gender-classification head to their MiE network, raising accuracy on CASME II to 75.20\% (vs. a single-task baseline of 72.36\%) and on SAMM to 55.88\% (vs. 51.47\%).
Wang et al. \cite{wang2023action} chose a more semantically related auxiliary task—AU detection—and obtained larger gains, reaching 86.34\% on CASME II and 81.28\% on SAMM. 
Despite these improvements, the margin over single-task models remains modest, suggesting that the limited number of MiE samples constrains how much multiple instance learning alone can enrich the learned representation.
A plausible explanation for the marginal gains observed in multi-task learning settings is the limited scale of existing MiE datasets. 
When training data for the primary task is scarce, the benefit of auxiliary supervision is inherently constrained.

Self-supervised motion representation for MiE recognition.
Fan et al. \cite{fan2023selfme}  proposed a self-supervised approach designed to learn motion symmetry features from optical flow.
The rationale behind this method is that symmetrical facial motion patterns are indicative of genuine expressions and are critical in distinguishing subtle MiEs. 
The approach reaches a UAR of 92.9\% on CASME II and 70.12\% on SMIC-HS. 
Wang et al. \cite{wang2023temporal} used self-supervised learning to pre-train the model on a new interpolation dataset to capture subtle and fast facial movements.
The method achieved advanced performance, with UAR scores of 92.71\% on CASME II and 74.04\% on SAMM.
While Self-supervised learning leverages temporal information and motion continuity to improve MiE recognition, they remain susceptible to noise, especially due to the low amplitude and transient nature of MiEs.

Lightweight deep learning seeks to curb over-fitting and reduce computational overhead by designing highly compact architectures—an advantage for latency-sensitive or edge deployments. Khor et al. \cite{khor2019dual} pruned AlexNet and used the truncated network as a dual-stream backbone. By shortening several convolutional layers, they cut the parameter count from 62.38 M to 0.63 M while still boosting accuracy on CASME II and SMIC by 71.19\% and 63.41\%, respectively (vs. 62.96\% and 59.76\%). Liong et al. \cite{liong2019shallow} further reduced complexity through two strategies: (i) limiting the input to optical-flow information between onset and apex frames, and (ii) employing a depth-2 3D CNN backbone. Their model contains only 0.0016 M parameters yet attains competitive UAR scores of 83.82\% on CASME II and 65.88\% on SAMM. Overall, lightweight architectures can markedly shrink model size and inference latency while maintaining, and in some cases improving, recognition performance—making them well-suited for MiE recognition on resource-constrained devices.

Other learning paradigms have also been explored to relieve the practical bottlenecks of MiE recognition. 
Because existing MiE datasets are small and collected under heterogeneous recording protocols, models trained on one corpus rarely transfer well to new scenarios. Cross-domain few-shot learning tackles this issue by importing knowledge from source domains with different annotation schemes and feature distributions.
Dai et al. \cite{dai2021cross} combined metric-based few-shot learning with AU guidance, boosting accuracy from 57.63\% to 81.88\% in a small domain shift setting (CASME II to CASME) and from 48.03\% to 65.12\% under a large domain shift (CASME II to SMIC). 
Gong et al. \cite{gong2023meta} embed MaE and MiE features into a shared metric-learning space, achieving 80.95\% and 63.13\% accuracy on CASME II and SMIC, respectively.
Although cross-domain few-shot learning and meta-learning alleviate data scarcity, their performance remains sensitive to domain discrepancy and class imbalance. 
GAN provides a way to increase data samples by generating synthetic MiE images.
Liong et al. \cite{liong2020evaluation} first used a GAN to produce optical-flow maps for data augmentation and reported 73.01\% accuracy on both the SMIC and CASME II datasets. 
Subsequent studies directly generate synthetic MiE frames \cite{li2021improved,yu2021ice,zhou2023ulme,10375342}. 
Most of these approaches incorporate AU priors to guide the generator, yielding finer-grained and temporally smoother facial motions and lifting recognition performance to 80.0\% on SAMM and 70.8\% on MMEW.
GAN-based augmentation remains sensitive to AU-annotation accuracy, suffers from training instability, and may introduce distributional artifacts that do not fully match real MiE data. As a result, models trained on heavily synthesized samples can overfit generator biases, and the true generalizability of such systems in the wild still needs systematic validation.

\section{In-Depth Discussions on Challenges and Future Directions}
\textcolor{black}{Despite its current prosperity of facial expression analysis in various domains, there remain several challenges and issues that require attention in the future. Hereinafter, we provide five notable research directions.}

\subsection{Advancing Facial Expression Datasets}
\textcolor{black}{Current datasets for training facial expression analysis systems face significant challenges. Datasets for MiE are particularly limited due to their subtle and transient nature, making collection and annotation difficult. Most MiE datasets are collected in controlled laboratory settings, restricting their applicability in real-world scenarios. Variations in methodologies and paradigms for data collection further complicate standardization and scalability.
Annotating MiE data requires certified expertise, significantly increasing the time and resources needed for dataset creation. In contrast, MaE datasets are generally larger but face challenges such as biases and inconsistencies. Many MaE datasets are derived from publicly available sources, leading to varying data quality and potential bias.
To overcome these challenges, future research should focus on three key areas:
\textbf{1). Unsupervised and semi-supervised learning:} reducing dependence on large annotated datasets by enabling models to learn from limited or unlabeled data; 
\textbf{2). Synthetic data generation:} using GANs and diffusion models to create realistic facial expression samples that preserve analytical features; 
\textbf{3). Standardized protocols for data collection:} developing universal standards for consistent, interoperable datasets. For example, designing collection protocols for ``wild'' data to capture diverse, naturalistic expressions via IoT devices or ubiquitous cameras.
}

\subsection{Adaptation to Diverse Cultural Contexts}
\textcolor{black}{
Facial expression analysis systems often presuppose universal emotional displays, yet cultural norms strongly shape both the production and interpretation of emotion.
MiEs are regarded as cross‑culturally consistent signals of concealed emotion, but their frequency and visibility still vary among populations. 
In contrast, MaEs are explicitly molded by culturally learned display rules that dictate when, where, and how emotions may be shown.
Cultural-adaptive facial expression analysis therefore requires datasets that span a broad spectrum of regions, age groups, and social contexts, capturing both the fleeting subtleties of MiEs and the intensity‑and‑duration patterns of culture‑specific MaEs. 
Contextual metadata—interaction setting, social hierarchy, local norms—should accompany each sample to sharpen model relevance.
Global IoT deployments (e.g., smart assistants, surveillance systems, wearables) amplify this need. 
Transfer‑learning pipelines that fine‑tune pretrained networks on region‑specific datasets can localize performance without costly full retraining. 
Moreover, contextual fusion—combining facial cues with environmental signals such as social setting or cultural events—enables devices to deliver feedback that is both accurate and culturally appropriate.
By pairing culturally diverse data with context‑aware modeling, facial expression analysis systems can achieve higher accuracy and broader applicability, ensuring reliable operation across heterogeneous cultural environments.
}

	\subsection{Data Privacy and Protection}
		\textcolor{black}{
        Data privacy and security are critical challenges in integrating facial expression analysis into IoT systems since facial data often contains sensitive biometric information. 
		The collection, transmission, and storage of data from distributed IoT devices (e.g., smart home cameras) introduce significant privacy concerns. 
		Many IoT-derived datasets lack proper regulation, and the reliance on wireless communication protocols increases vulnerability to interception and hacking. 
		Weak encryption and unsecured channels can expose sensitive facial data during transmission. 
		Centralized storage systems and cloud-based solutions heighten the risk of data breaches, compounded by the absence of a unified global privacy framework. Although regulations (e.g., the European Union General Data Protection Regulation) set standards for biometric data protection, enforcement varies, complicating global compliance.
		Synthetic facial expression data offers a promising solution to privacy concerns. 
		Generative models (e.g., GANs) and advanced text-to-video systems (e.g., Sora) can create realistic synthetic data to supplement real datasets and enhance model performance while safeguarding privacy.
		Federated learning further enhances privacy by enabling decentralized model training, where sensitive data remains on user terminals. 
		This method minimizes risks of unauthorized access and data leakage, as only aggregated model updates are shared with a central server. 
		By preventing raw data transmission, federated learning can preserve individual privacy while supporting collaborative training. 
		However, federated learning on resource-constrained IoT devices remains a challenge; therefore, it is necessary to develop of efficient algorithms that balance performance, privacy, and energy consumption. 
		Future research should focus on optimizing these techniques to enable secure, scalable, and efficient facial expression analysis in IoT systems.
		}

	\subsection{Real-Time Processing}
\textcolor{black}{
Real‑time processing is imperative for facial expression analysis in IoT scenarios (e.g., emotion‑aware assistants and healthcare monitors), but delivering efficient, scalable performance is hampered by algorithmic complexity, tight power and memory budgets, and dynamic operating conditions.
Facial expression analysis comprises compute‑heavy stages (feature extraction, expression classification, temporal modeling) that typically rely on deep networks with millions of parameters, while continuous video streaming further taxes processing cycles, memory, and battery life.
Latency is an additional constraint: cloud inference incurs transmission delays and network congestion, unacceptable in time-critical settings like medical monitoring.
Shifting computation to the edge alleviates these bottlenecks by processing data locally and conserving bandwidth, especially when paired with on‑device accelerators (e.g., GPUs, or custom AI chips). 
Model‑compression techniques—including quantization, pruning, and knowledge distillation—further shrink computational load while retaining accuracy, making deep models deployable on constrained hardware.
Neuromorphic cameras or motion-triggered inference have the potential to reduce energy consumption by activating models only when expression-relevant motion is detected.
Lean communication protocols are essential to sustain real‑time throughput over bandwidth‑limited IoT links. 
Addressing these challenges holistically will unlock robust, low‑latency facial expression analysis systems tailored to IoT environments.
}

	\subsection{Comprehensive Analysis}
    \textcolor{black}{
	Comprehensive facial expression analysis involves multi-modal approaches that extend beyond the traditional focus on facial muscle movements. 
	Recent studies have introduced various modalities such as electromyography \cite{kim2022classification}, word \cite{wahid2023human}, electroencephalogram \cite{wang2023micro,zhao2024micro} and other psychological signals \cite{saffaryazdi2022using} can also reveal the emotions. 
	Developing a multi-modal analysis of facial expressions could take advantage of complementary information and obtain robust features of emotions. 
	Furthermore, several datasets combined with physiological signals have been proposed. 
	For example, Chen et al. \cite{chen2019analyze} introduced the micro-gesture dataset for hidden emotion recognition. 
	Zou et al. \cite{zou2022concordance} proposed a facial expression dataset with synchronized physiological signals. 
	More valuable research can explore the intrinsic relationship between facial expressions and other signals in the future.
	In addition to applying multi-modal expression analysis methods to IoT, IoT needs to consider multi-tasks for collaborative analysis.
	In the context of the IoT, multi-modal expression analysis should also address the challenges of multi-tasking. 
	IoT devices often need to process data from diverse sensors and perform various analyses simultaneously. 
	This necessitates the development of efficient multi-task learning models to enhance system performance and responsiveness. 
	For example, a smart home system might simultaneously analyze data from security cameras, environmental sensors, and facial expressions to enable collaborative decision-making.
	Future directions for facial expression analysis in IoT should prioritize the integration of multi-modal and multi-task capabilities. 
	By incorporating various physiological signals, researchers can develop more accurate and comprehensive emotion recognition systems. 
	In addition, by improving the efficiency of multi-task learning models, IoT devices can maintain high performance and responsiveness while executing complex tasks. 
	This will enable IoT systems to better understand and respond to emotional demands of users to provide more intelligent and human-centric services.
	}

\end{document}